\documentclass{article}

\usepackage{PRIMEarxiv}

\usepackage[utf8]{inputenc} 
\usepackage[T1]{fontenc}    
\usepackage{hyperref}       
\usepackage{url}            
\usepackage{booktabs}       
\usepackage{amsfonts}       
\usepackage{nicefrac}       
\usepackage{microtype}      
\usepackage{lipsum}
\usepackage{fancyhdr}       
\usepackage{graphicx}       
\graphicspath{{media/}}     

\usepackage{epsfig}
\usepackage{graphicx}
\usepackage{amsmath}
\usepackage{amssymb}
\usepackage{soul}
\usepackage{colortbl}
\usepackage{comment}
\usepackage{multirow}
\usepackage{diagbox}
\usepackage{slashbox}
\usepackage{float}
\usepackage[dvipsnames]{xcolor}

\usepackage{makecell}
\usepackage{verbatim}
\usepackage{subfig}

\title{Large Language Models Still Face Challenges in Multi-Hop Reasoning with External Knowledge}

\author{
  Haotong Zhang \\
  Department of Computer Science\\
  The University of Manchester\\
  \texttt{haotong.zhang@manchester.ac.uk} \\
}

\begin{document}
\maketitle

\begin{abstract}
We carry out a series of experiments to test large language models' multi-hop reasoning ability from three aspects: selecting and combining external knowledge, dealing with non-sequential reasoning tasks and generalising to data samples with larger numbers of hops. We test the GPT-3.5 model on four reasoning benchmarks with Chain-of-Thought prompting (and its variations). Our results reveal that despite the amazing performance achieved by large language models on various reasoning tasks, models still suffer from severe drawbacks which shows a large gap with humans.
\end{abstract}

\section{Introduction}
``Reasoning'' is the process of drawing conclusions from existing information, with the aim of seeking the truth. It is often considered a unique human ability. How to endow artificial intelligence systems with this ability is one of the principal concerns in Natural Language Processing (NLP). 

The emergence of language models revolutionised the landscape of NLP \cite{devlin2018, radford2019, yang2019, raffel2020, he2023}. Pretraining on large text corpora enables models to have a preliminary grasp of natural language. Therefore, models can adapt to downstream tasks easily with some finetuning.

Scaling up the scale of language models further boosts the performance. In the last several years, with the rapid growth of the pretraining data and a significant increase in computing power, we have witnessed a dramatic improvement in large language models' (LLMs) performance on a variety of NLP tasks, even without finetuning \cite{brown2020, chowdhery2023, openai2024}.

However, these models still fall short of challenging reasoning tasks. Chain-of-Thought prompting and follow-up approaches make up for this gap \cite{wei2022cot, zhou2023, yao2023tot}. Under the guidance to decompose problems, models demonstrate amazing ability on reasoning tasks, even surpassing the performance of state-of-the-art supervised models and humans.

In this paper, we focus on multi-hop reasoning tasks in natural language with external knowledge. Formally speaking, defining the task of \emph{multi-hop reasoning} is not straightforward, since the definition of a \emph{hop} is unclear. \cite{mavi2022} defines multi-hop reasoning as the reasoning task where the number of required \emph{contexts} (\emph{e.g.}, a sentence, a paragraph, an entity in a knowledge graph) to solve each problem should be strictly larger than 1, to distinguish from \emph{single-hop reasoning}. The number of required contexts is the number of hops. 

LLMs have encountered massive text corpus during pretraining and learnt knowledge from them. Such knowledge is stored in the parameters of LLMs and is called \emph{internal knowledge}. In contrast, additional information provided to LLMs in downstream tasks is called \emph{external knowledge}.

Notwithstanding all the amazing performance observed on LLMs, we argue that they are still inadequate to completely solve the multi-hop reasoning tasks with external knowledge, even with Chain-of-Thought prompting. We conduct experiments in each part of the reasoning process: 
1) \textbf{Selecting and combining with external knowledge}. We compare model performance with and without external knowledge (\emph{i.e.}, only with internal knowledge) and analyse how the proportion of distractors influences model performance. We also provide counterfactual knowledge to the model to see how the model deals with confusion caused by knowledge inconsistency; 
2) \textbf{Decomposing problems and generating reasoning paths}. We compare the model's ability to decompose problems and generate reasoning paths on sequential and non-sequential reasoning samples; 
3) \textbf{Generalising to samples with larger numbers of hops}. We observe whether providing samples with larger numbers of hops in the prompt boosts the model's generalisation ability.

\section{Related Work}
\subsection{Multi-hop Reasoning}

Multi-hop reasoning sometimes involves a status transition of entities via relations, especially when the provided context is a knowledge graph. To this end, graph neural networks (GNNs) are commonly used to leverage edges (\emph{e.g.}, entity-entity relations) to promote information aggregation and integration between nodes (\emph{e.g.}, entity information) \cite{staliūnaitė2022, zhang2022, du2022, cui2023, hemmati2023}.

Many multi-hop reasoning tasks come with accompanied contexts, which aid the reasoning process. To this end, a mechanism called \emph{Retriever-Reader} splits the task into two parts: 
1) A \emph{retriever} chooses relevant knowledge in the contexts using a query, which is updated each step; 
2) A \emph{reader} generates the answer based on the chosen knowledge along with the question \cite{xiong2021, zhao2021, wu2022}.

Since existing models perform quite well on single-hop reasoning \cite{xu2021, wang2021}, some approaches decompose multi-hop questions into a set of single-hop subquestions and answer them one by one \cite{min2019, press2023, zhang2024}.

\subsection{Reasoning with Large Language Models}

Large language models often have a huge number of parameters (usually over 100 billion). Their large scale brings about steady improvement in reasoning tasks over small language models, which involve fundamental aspects of human intelligence \cite{touvron2023, bang2023}. 

On the other hand, LLMs' large scale causes finetuning to be extremely expensive. Thanks to the effective pretraining process, LLMs' internal knowledge enables them to perform well on unseen tasks (\emph{zero-shot learning}) \cite{wei2022, sanh2022}. \emph{Few-shot learning} via \emph{prompting} further raises LLMs performance by helping them to quickly adapt to specific downstream tasks. The prompt includes a few examples (\emph{exemplars}), which are in the desired format of the downstream task, to instruct the model what to do (\emph{e.g.}, what is the requirement of the task, what should be output). Remarkably, this method has been demonstrated successful for a range of tasks \cite{brown2020}. 

\subsection{Chain-of-Thought Prompting}

Chain-of-Thought (CoT) prompting is a few-shot mechanism on LLMs \cite{wei2022cot}. It guides the model to decompose a multi-hop problem into intermediate steps and solve each one before giving the final answer. CoT leads to a huge improvement in the model performance on many reasoning tasks.

Later, the following studies improve CoT prompting from different aspects. \cite{wang2023} propose self-consistency: instead of generating one reasoning path, the model is required to generate several reasoning paths and choose the answer supported by most paths (\emph{i.e.}, majority voting). \cite{yao2023tot} expand the left-to-right reasoning path into a tree-like path and plug in simple search algorithms. Although the tree-like structure allows backtracking and lookahead, it still modifies within a sequential reasoning path, but cannot handle non-sequential cases. \cite{zhou2023} modify the original CoT prompt by adding an explicit decomposition stage. This modification substantially improves the model's reasoning and generalisation ability on samples with larger numbers of hops. \cite{kojima2022} test the zero-shot ability of LLMs. Instead of giving a few exemplars as prompts to encourage the model to decompose problems, they just add ``Let's think step by step'' before every question. This simple method even surpasses some standard few-shot methods. \cite{creswell2023} and \cite{yao2023react} explore the ability of LLMs to combine with external knowledge since some reasoning tasks require specific knowledge that might not be frequently encountered during pretraining. External knowledge provides accurate information to LLMs and reduces hallucinations.

In this paper, we conduct experiments following the idea of CoT prompting (and its variations). Our results show that LLMs fail to maintain good performance under various complex situations, which reveals the hidden weakness of current LLMs.

\section{Experimental Setup}

\begin{table*}[!ht]
\centering
\centerline{
\begin{tabular}{c|c}
\hline
\textbf{Benchmark} & \textbf{Example} \\
\hline
\textbf{HotpotQA} & \makecell[l]{\textbf{\emph{Context}}: Scott Derrickson is an American director, screenwriter and producer. Edward Davis \\ Wood Jr. was an American filmmaker, actor, writer, producer, and director.\\ \textbf{\emph{Question}}: Were Scott Derrickson and Ed Wood of the same nationality? \\ \textbf{\emph{Answer}}: Yes.} \\
\hline
\textbf{EntailmentBank} & \makecell[l]{\textbf{\emph{Context}}: Scraping an object may cause small particles to break off of that object. Very small \\ rocks / minerals are kinds of of small particles. Glacier may scrape rocks when it moves. A \\ rock is a kind of object. Soil is made of very small rocks / minerals. \\ \textbf{\emph{Question}}: A glacier is a slow moving river of ice. How does a glacier help create soil? \\ \textbf{\emph{Answer}}: It scrapes small particles off large rocks.} \\
\hline
\textbf{QASC} & \makecell[l]{\textbf{\emph{Context}}: All cnidarians are aquatic. Cnidarians include jellyfish and anemones. \\ \textbf{\emph{Question}}: What are aquatic animals? (A) Pelycosaur (B) candy (C) water (D) angiosperm \\ (E) weater (F) jellyfish (G) arachnids (H) cookies. \\ \textbf{\emph{Answer}}: F.} \\
\hline
\textbf{bAbI15} & \makecell[l]{\textbf{\emph{Context}}: Mice are afraid of cats. Sheep are afraid of wolves. Jessica is a lion. Emily is a wolf. \\ Gertrude is a sheep. Winona is a cat. Deer are afraid of lions. Snakes are afraid of eagles. \\ \textbf{\emph{Question}}: What is Gertrude afraid of? \\ \textbf{\emph{Answer}}: Wolves. \\ \textbf{\emph{Question}}: What is Jessica afraid of? \\ \textbf{\emph{Answer}}: Not mentioned.} \\
\hline
\end{tabular}}
\caption{Sample example from each benchmark.}
\label{tab:data}
\end{table*}

We carry out our experiments on four benchmarks in natural language reasoning.
\textbf{HotpotQA} \cite{yang2018} is a two-hop question-answering dataset of which each sample is accompanied by several Wikipedia paragraphs. 
\textbf{EntailmentBank} \cite{dalvi2021} is a multi-hop entailment dataset on elementary science. The number of hops ranges from 1 to 17. Some samples have non-sequential reasoning paths. We modify it slightly so that samples can fit in the question-answering form.
\textbf{QASC} \cite{khot2020} is a two-hop reasoning dataset on elementary science in the form of eight-way multiple-choice question-answering. 
\textbf{bAbI15} \cite{weston2015} is a two-hop deduction question-answering dataset. We modify it slightly so that 50\% of the samples are \emph{non-deductive}. Specifically, there is not enough information in the contexts to answer such questions, and the model should output something like \emph{Not mentioned}. 
An example of each benchmark is shown in Table~\ref{tab:data}. 

Unless otherwise stated, all experiments are carried out on text-davinci-002 and follow the Chain-of-Thought method. In each experiment, we give four exemplars in the prompt and construct a test set with 100 samples. Prompts used for each experiment are shown in the Appendix.

\section{Internal versus External Knowledge}

\subsection{Internal Knowledge Is Not Enough}

As a consequence of the massive text corpus seen during pretraining, it is generally believed that LLMs themselves contain enough implicit knowledge to complete the reasoning tasks. CoT prompts further motivate the models' ability to use implicit knowledge. However, experiments on HotpotQA and EntailmentBank show that the model performance is still unsatisfactory. The model is forced to rely on its internal knowledge to answer the question as no external information is provided. As shown in the last column of Table~\ref{tab:hotpotqa1} and Table~\ref{tab:entailmentbank1}, the no-context setting, accuracy is quite low\footnote{We check each sample manually and consider the answer as correct if it has the same meaning as the ground truth.}.

Two benchmarks suffer from different issues. In HotpotQA, the model tends to generate wrong knowledge out of thin air. This is probably because HotpotQA requires much specific knowledge which might not be frequently encountered during pretraining. In EntailmentBank, however, using irrelevant knowledge (\emph{i.e.}, correct but not helpful to answer the question) takes the primary responsibility for the poor performance. We observe that in many samples of EntailmentBank, the model just retrieves knowledge from its memory based on the keyword in the question, and randomly chooses common words among this knowledge as the answer, without any reasoning process. 

Furthermore, though these two benchmarks were constructed before the model's pretraining began, we would like to point out another common phenomenon that many LLMs suffer from: their implicit knowledge is not up-to-date. Due to the computationally intensive pretraining process and the rapid update of knowledge in the real world, LLMs are almost blind to the new information that emerged after their pretraining process stopped. 

We argue that external knowledge supplies LLMs with additional information which might not be captured during pretraining, and makes sure that the knowledge is accurate and up-to-date. Due to the input limit of LLMs, a common way is to provide each sample with corresponding knowledge.

\begin{table}
\centering
\begin{tabular}{l|rrrr}
\hline
{} & \textbf{100\%} & \textbf{50\%} & \textbf{20\%} & \textbf{NC} \\
\hline
\textbf{EM} & 100.00 & 76.00 & 66.00 & N/A \\
\textbf{OneHit} & 100.00 & 93.00 & 91.00 & N/A \\
\textbf{Accuracy} & 83.00 & 74.00 & 73.00 & 40.00 \\
\hline
\end{tabular}
\caption{EM, OneHit and accuracy on HotpotQA in different settings. NC means no-context.}
\label{tab:hotpotqa1}
\end{table}

\begin{table}
\centering
\begin{tabular}{l|rrrr}
\hline
{} & \textbf{100\%} & \textbf{50\%} & \textbf{20\%} & \textbf{NC} \\
\hline
\textbf{Precision} & 95.22 & 64.76 & 50.84 & N/A \\
\textbf{Recall} & 95.60 & 84.05 & 77.62 & N/A \\
\textbf{Accuracy} & 77.00 & 68.00 & 59.00 & 35.00 \\
\hline
\end{tabular}
\caption{Precision, recall and accuracy on EntailmentBank in different settings. NC means no-context.}
\label{tab:entailmentbank1}
\end{table}

\subsection{Adding and Selecting External Knowledge}

\begin{figure*}
\begin{center}
\includegraphics[width=1\linewidth]{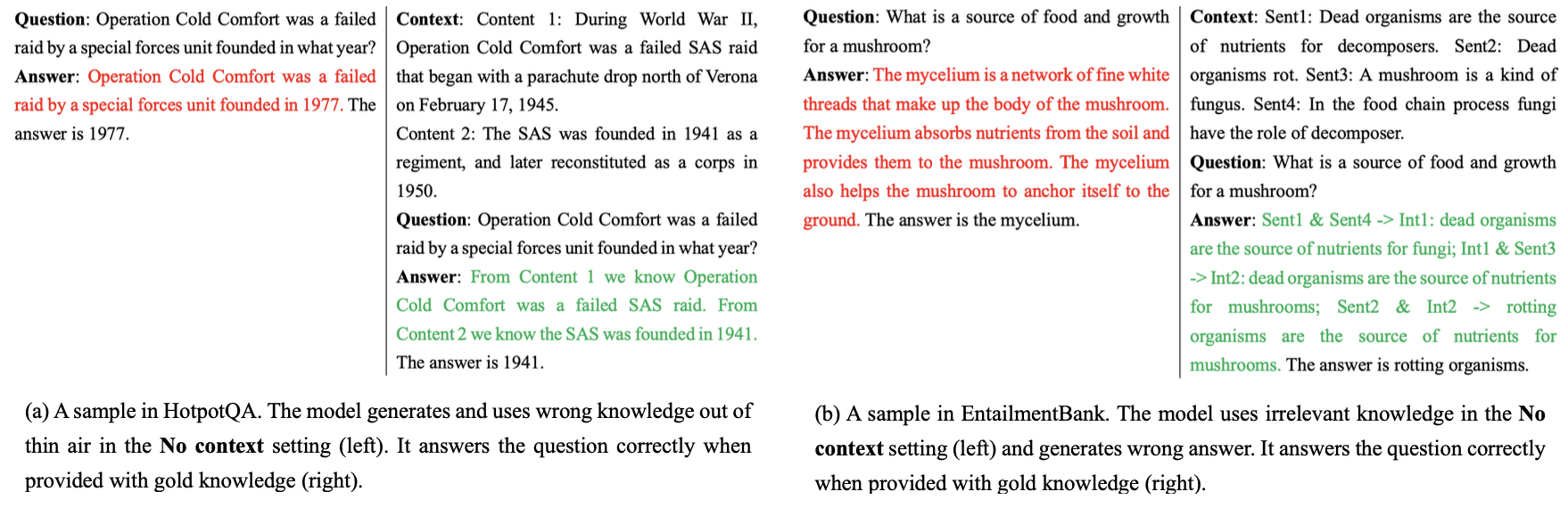}
\end{center}
\caption{Two examples on HotpotQA and EntailmentBank respectively where providing external knowledge helps the model answer the question correctly. Best viewed in colour.}
\label{fig:examples}
\end{figure*}

We use the same exemplars and test sets as in the last experiment but add external knowledge to each sample to observe the effects. We also include distractors (\emph{i.e.}, irrelevant knowledge) along with gold knowledge (\emph{i.e.}, knowledge used to do reasoning) and control the proportion of gold knowledge to distractors in each setting. Results are shown in the first three columns in Table~\ref{tab:hotpotqa1} and Table~\ref{tab:entailmentbank1}\footnote{Results shown here are generated by the standard CoT prompting. Since the two tasks both involve the knowledge selection step, we have also tried the Selection-Inference prompting \cite{creswell2023}. However, it performs much worse than standard CoT. We suspect that the performance difference between ours and theirs is due to the different data used. In their experiments, samples follow a similar format, so the model can do reasoning quite well in the inference step even without access to the question. However, in our experiments, there is no obvious pattern. The model cannot even select the gold knowledge, let alone make correct inferences.}. Column headings mean 100\%, 50\% and 20\% of the knowledge in the provided context is gold knowledge, respectively. Since each sample in HotpotQA is two-hop (requires two pieces of knowledge to answer), we use exact match (EM) and \emph{OneHit} to measure the model's ability to select gold knowledge. EM reflects whether the model correctly finds both gold knowledge (we ignore the sequence of knowledge). OneHit computes whether the model can find at least one piece of gold knowledge. On the other hand, we use precision and recall because the numbers of hops of samples in EntailmentBank are different. Precision is the proportion of selected gold knowledge over all selected knowledge. Recall is the proportion of selected gold knowledge over all gold knowledge. Accuracy in both tables measures the answer accuracy.

It is obvious that additionally including external knowledge boosts model performance, regardless of the presence of distractors. The results are quite intuitive on HotpotQA as provided contexts make up for the lack of specific knowledge of the model. It is somewhat surprising though, that on EntailmentBank, both with-distractor settings outperform the no-context setting a lot even though there is irrelevant knowledge provided. We speculate that this is because explicitly providing gold knowledge in the contexts increases its probability of being chosen by the model. Figure~\ref{fig:examples} shows examples where the model fails in the no-context setting but answers correctly when contexts are provided.

In HotpotQA, we notice that as the proportion of gold knowledge decreases, EM decreases the fastest while OneHit decreases the slowest, and accuracy is between them. It is intuitive since adding more distractors makes it more difficult for the model to select all gold knowledge, but is still quite easy to hit at least one. It also implicitly indicates that there are shortcuts in some samples where the question can be answered by only one piece of gold knowledge. Therefore, we observe a smaller decline in accuracy compared to EM. The existence of shortcuts can be further backed up by the fact that the accuracy is higher than EM in the 20\% setting, which means that the model correctly answers some samples even with the wrong gold knowledge. Taking this factor into consideration, the results are still quite reasonable since accuracy is always lower than OneHit in each setting.

In EntailmentBank, precision drops more rapidly than recall and accuracy with the proportion of gold knowledge decreasing. High recall but low precision, when distractors are introduced, demonstrates that the model selects most of the gold knowledge but also many irrelevant ones. As we have found out in case studies, this result is caused by the model selecting knowledge based on overlapping keywords. Although overlapping words are a good indicator of gold knowledge, they can also mistakenly include irrelevant knowledge. 

In both benchmarks, we observe hallucinations during the reasoning process. The model generates something that is not from the given contexts. The phenomenon is more severe in EntailmentBank, which requires a lot of elementary science knowledge that can be easily obtained from pretraining. This also explains why the precision in the 100\% setting in Table~\ref{tab:entailmentbank1} is less than 100\%: the model uses some knowledge not provided. We witness a larger and faster accuracy decline in EntailmentBank than in HotpotQA when there are more distractors. We suppose that more complex reasoning tasks are more sensitive to distractors.

In both benchmarks, adding distractors indeed damages the model's ability to select gold knowledge. However, the proportion of distractors does not matter so much, as the gap between 50\% and 20\% settings is always smaller than the gap between 100\% and 50\% settings on all metrics. We conjecture that although LLMs may not be able to select the gold knowledge, they have a higher probability to select the most relevant knowledge than the less relevant one. Therefore, as the proportion of distractors increases, the degradation of model performance will decrease.

In reality, sometimes it is difficult to provide corresponding contexts for each sample. Instead, a large knowledge base with millions of pieces of knowledge is provided to aid the reasoning task. Due to the input limitation of LLMs, we cannot do experiments in this setting, however, we can infer its performance based on the trend shown in Table~\ref{tab:hotpotqa1} and Table~\ref{tab:entailmentbank1}. Providing a large knowledge base is equivalent to the with-distractor setting where the proportion of gold knowledge is very small. The model performance will surely be lower than the 20\% setting, but might not drop too much if a good retrieval mechanism is used to filter out most irrelevant knowledge. However, the performance will still be much lower than human performance\footnote{83.6\% on HotpotQA. The authors of EntailmentBank did not publish human performance.}. It is worth thinking about how to integrate the knowledge base into LLMs.

\subsection{Counterfactual Knowledge}

Counterfactual reasoning is a special kind of reasoning which enables the capacity to shift from perceiving the immediate environment to an alternative, imagined perspective \cite{van2015}. Sometimes, the task requires reasoning on something different from the real world (\emph{e.g.}, imagined past events or future outcomes not yet at hand) to learn from experience or enable planning and prediction. However, following experiments show that providing knowledge inconsistent with the model's internal knowledge causes great confusion.

Experiments are carried out on QASC and bAbI15. We provide counterfactual knowledge (\emph{i.e.}, knowledge that is false) to some samples. In QASC, the correct answer always occurs in the context. We substitute it with a wrong choice provided in the question (so the context becomes counterfactual), thus changing the answer. In bAbI15, we swap the subject and object in all factual sentences. For example, \emph{Mice are afraid of cats} becomes \emph{Cats are afraid of mice}. The knowledge provided in the prompt and the test set is either factual or counterfactual, yielding four different settings. 

In QASC (Table~\ref{tab:qasc}), we provide the overall accuracy in each setting, and the \emph{pseudo accuracy} when the provided knowledge is counterfactual in the test set. Pseudo accuracy means the answer is the same as the original one (\emph{i.e.}, the ground truth for the corresponding sample whose context is factual). In bAbI15 (Table~\ref{tab:babi15}), there are two values in each cell. The overall accuracy reveals model performance on all samples. The \emph{non-deductive accuracy} only counts the accuracy on non-deductive samples. Since the model answers nearly all deductive samples correctly, the second accuracy provides more information on the model's real performance.

\begin{table}
\centering
\begin{tabular}{l|rr}
\hline
{} & \textbf{Factual} & \textbf{Counterfactual} \\
\hline
\textbf{Factual} & 97 & 91 \\
\textbf{Counterfactual} & 75/20 & 74/19 \\
\hline
\end{tabular}
\caption{Overall accuracy/Pseudo accuracy (for the second row only) in each setting on QASC. Column headings and row headings indicate the factualness of the provided knowledge in the prompt and the test set.}
\label{tab:qasc}
\end{table}

\begin{table}
\centering
\begin{tabular}{l|rr}
\hline
{} & \textbf{Factual} & \textbf{Counterfactual} \\
\hline
\textbf{Factual} & 61/11 & 49/2 \\
\textbf{Counterfactual} & 48/0 & 50/1 \\
\hline
\end{tabular}
\caption{Overall accuracy/Non-deductive accuracy in each setting on bAbI15. We provide the number of correct samples (out of 50) instead of the accuracy percentage for the second value, so it is more straightforward to compute the number of correct deductive samples from these two values. Column headings and row headings indicate the factualness of the provided knowledge in the prompt and the test set.}
\label{tab:babi15}
\end{table}

In the QASC task, the factual-prompt-factual-test setting (\emph{i.e.}, the top-left cell) achieves the highest overall accuracy, which is consistent with our intuition. The overall accuracy is higher in the first row than in the second row, and also higher in the first column than in the second column. This means that the model performs better when factual knowledge is provided in at least one of the prompt or the test set. The gap between the overall accuracy is larger between two rows than between two columns, which reveals that the factualness of provided knowledge in the test set influences model performance more. We argue that this is probably because the inconsistency between knowledge in the test set and the knowledge encountered during pretraining confuses the model. The evidence is that the two values in each cell of the second row of Table~\ref{tab:qasc} add up to approximately the overall accuracy in the first row. A deeper investigation into each sample where the model generates the original answer (\emph{i.e.}, the ground truth before the context is modified) reveals that, in all such samples, the model generates the factual knowledge out of thin air, and follows it to derive at the answer. This means that the model can generate the correct reasoning path, but it uses the wrong knowledge. On the other hand, since the modification of knowledge mostly relates to entity replacement, the model can still apply the reasoning format learnt from the prompt smoothly on the test set. Therefore, the knowledge factualness in the test set guarantees a good performance. 

In contrast, although we still witness the highest overall accuracy in the factual-prompt-factual-test setting in the bAbI15 task, a simple calculation shows that there is almost no difference in model performance on deductive samples. The model correctly answers nearly all deductive samples in all settings, showing that the factualness of provided knowledge has little influence on them. However, the model fails on most non-deductive samples, even in the factual-prompt-factual-test setting. 

An observation of non-deductive samples reveals that there are two main reasons for the poor performance: 
1) \textbf{Word bias}. The model fails some non-deductive samples due to its inability to distinguish the subject and object. Instead of answering something like \emph{Not mentioned}, the model always outputs a specific answer. For example, if the question asks \emph{What is Emily afraid of?} based on the knowledge \emph{Emily is a wolf} and \emph{Sheep are afraid of wolves}, then the model will answer \emph{Emily is afraid of sheep}. This indicates that the model cannot understand the meaning of \emph{afraid}, but just depends on some word bias (\emph{i.e.}, one word is always associated with another, such as wolf and sheep). 
2) \textbf{Hallucination}. The most severe hallucination occurs when the knowledge provided in the prompt and the test set is inconsistent. Specifically, the model uses the knowledge provided in the prompt when doing reasoning on the test set. In the factual-prompt-counterfactual-test setting (\emph{i.e.}, the bottom-left cell), the model uses knowledge similar to \emph{Mice are afraid of cats} in 16\% of the non-deductive samples in the test set, where the provided knowledge is something like \emph{Cats are afraid of mice}. In the counterfactual-prompt-factual-test setting (\emph{i.e.}, the top-right cell), this proportion even increases to 72\%. The result is a bit surprising since the knowledge provided in the test set is consistent with the model's internal knowledge. However, the model still follows the knowledge seen in the prompt. It is clear that the model is influenced by the prompt significantly. A similar phenomenon is not observed in QASC, because knowledge provided in the prompt and the test set is unrelated. Even though the model answers some non-deductive samples correctly, it does not follow the format given in the prompt. Rather, the model generates something like \emph{Emily is a lion. Lions are not afraid of anything}, which is likely to be drawn from its internal knowledge. 

We conjecture that the different behaviour of the model on two tasks is due to two reasons. First, QASC involves elementary science knowledge that is commonly encountered during pretraining. So the model is more likely to follow its internal knowledge and ignores the provided context. Although the knowledge such as \emph{Sheep are afraid of wolves} provided in bAbI15 should be common in the pretraining data as well, there is also specific knowledge such as \emph{Emily is a sheep} which is only valid in the given context. Such a reading-comprehension-like task might force the model to stick to the provided knowledge. Second, there is always an answer to the samples in QASC, but half of the samples in bAbI15 are unanswerable (\emph{i.e.}, non-deductive samples). It is widely observed that LLMs are unlikely to answer \emph{I don't know} even to unanswerable questions, but always try to give an answer \cite{yin2023}. Therefore, the model borrows knowledge from the prompt to make unanswerable questions ``answerable''.

\section{Reasoning in Non-Sequential Cases}

Chain-of-Thought prompting boosts LLMs' performance on many reasoning tasks, including multi-hop reasoning. However, these tasks only involve sequential reasoning paths, where one step only depends on its previous one step, in a linear way. In the real world, reasoning paths could be much more complex (\emph{i.e.}, non-sequential).

An experiment is carried out on EntailmentBank. To make sure only samples with non-sequential reasoning paths are selected, we only choose those whose depth and length of the entailment tree are not the same. Each sample is only accompanied by gold knowledge, to get rid of the influence of distractors. The accuracy is 56\%, compared to the 78\% accuracy on samples with sequential reasoning paths. Obviously, the model does worse when the reasoning paths are more complex.

Though the model still manages to answer more than half of the samples correctly, none of the generated reasoning paths is correct. Specifically, the errors can be roughly divided into four types: 
1) \textbf{Omitting some gold knowledge}. The model does not include all gold knowledge in the reasoning path; 
2) \textbf{Hallucination}. The model uses knowledge out of thin air; 
3) \textbf{Repeating provided knowledge}. Most of the time, the generated intermediate conclusions are just a repetition of one of the provided knowledge. For example, if a reasoning step is \emph{sent1 \& sent2 -> int1}, then \emph{int1} is the same as \emph{sent1}; 
4) \textbf{Wrong intermediate conclusions/reasoning steps}. Some generated intermediate conclusions do not follow the reasoning step. For example, if a reasoning step is \emph{sent1 \& sent2 -> int1}, then \emph{int1} actually cannot be inferred from \emph{sent1} and \emph{sent2} (and sometimes we find that \emph{int1} is actually inferred from \emph{sent3}). 

In a word, even CoT prompts fall short of generating correct and coherent reasoning paths for non-sequential reasoning tasks. Instead of doing sound reasoning where one step follows its previous steps, it seems like the model is just imitating the formats provided in the prompt and trying to fit in the format without understanding the meaning. 

\section{Generalisation Ability}

\begin{table*}
\centering
\begin{tabular}{l|rrrrrrrrrrr}
\hline
{} & \textbf{1} & \textbf{2} & \textbf{3} & \textbf{4} & \textbf{5} & \textbf{6} & \textbf{7} & \textbf{8} & \textbf{9} & \textbf{10} & \textbf{Overall} \\
\hline
\textbf{1-hop} & 70 & 60 & 60 & 60 & 80 & 70 & 30 & 50 & 90 & 70 & 64 \\
\textbf{2-hop} & 100 & 40 & 30 & 50 & 60 & 20 & 20 & 40 & 60 & 40 & 46 \\
\textbf{3-hop} & 60 & 30 & 60 & 30 & 20 & 30 & 40 & 10 & 50 & 20 & 35 \\
\textbf{4-hop} & 50 & 40 & 0 & 10 & 60 & 20 & 20 & 40 & 40 & 20 & 30 \\
\hline
\end{tabular}
\caption{Accuracy breakdown on samples with different numbers of hops and overall accuracy in different settings on EntailmentBank. The column and row headings are the numbers of hops of the samples in the test set and the prompt, respectively.}
\label{tab:entailmentbank3}
\end{table*}

As \cite{jansen2017} points out, aggregating knowledge from several sources is challenging for current reasoning models due to \emph{semantic drift}, or the tendency for algorithms to quickly move off-topic when assembling long chains of knowledge. It is intuitive and observed that models generally perform worse on reasoning tasks requiring more steps. On the other hand, models tend to imitate the format of the prompt in few-shot learning \cite{min2022}. However, in the real world, it is difficult to include all exemplars with the exact number of hops we need in the prompt to guide the reasoning process. Therefore, the generalisation ability on the number of hops is very important.

An experiment is carried out on EntailmentBank. We use the Least-to-Most Prompting \cite{zhou2023}. Specifically, we have four settings where the number of hops of exemplars ranges from 1 to 4 in the prompt. In the test set, the number of hops ranges from 1 to 10, and the number of samples with different number of hops is the same. Results are shown in Table~\ref{tab:entailmentbank3}.

There is no obvious pattern in the accuracy breakdown based on samples with different numbers of hops, even for samples that are 1- to 4-hop (which means providing exemplars of $n$-hop does not necessarily enhance the model performance on $n$-hop samples). The overall accuracy decreases when the number of hops of exemplars increases. By checking the generated reasoning paths of samples with different numbers of hops, we find that there are two main types of errors: 
1) \textbf{Over-decomposition}. The model tends to think further than the question requires (\emph{i.e.}, the answer to one of the subquestions is the correct answer to the original question), thus giving the wrong answer. 
2) \textbf{Wrong decomposition}. The model generates irrelevant subquestions. 
In both types, it seems like the model does not truly understand the question (or the meaning of \emph{decomposition}), but just tries to generate some subquestions. In fact, we find that even in the samples where the model answers correctly, the decomposition is a total mess. Requiring the model to explain why it derives the answer leads to a drop in performance. This phenomenon contradicts that in \cite{zhou2023}, where the models can generate clear and good reasoning paths on many tasks and promote performance. This might still be due to the complexity of reasoning paths. Sequential ones are quite easy and straightforward to LLMs, but non-sequential ones confuse them.

\begin{figure}
\begin{center}
\includegraphics[width=1\linewidth]{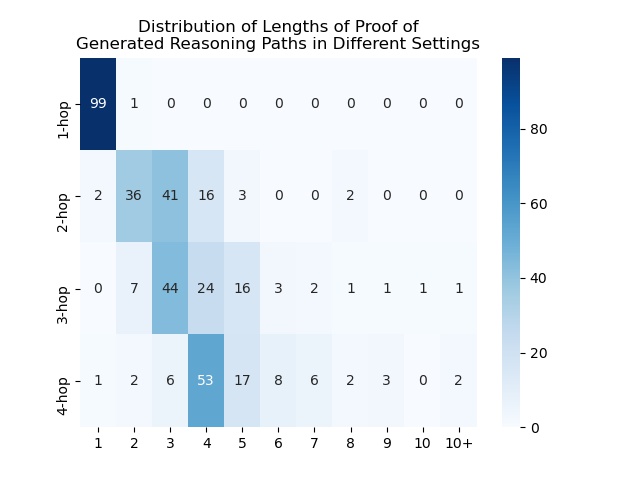}
\end{center}
\caption{Distribution of length of proof of generated reasoning paths in different settings. X-axis indicates the lengths of proof. Y-axis indicates different settings.}
\label{fig:entailmentbank}
\end{figure}

We further analyse the lengths of generated reasoning paths of samples in the test set in different settings, as shown in Figure~\ref{fig:entailmentbank}. The darker part moves to the right as the number of hops of exemplars increases. Except for the 1-hop setting, the model tends to generate reasoning paths of length from $n$ to $n+2$ in the $n$-hop setting. This indicates that the number of hops of exemplars does influence the model behaviour. Furthermore, in the 1-hop setting, though the model achieves the highest overall accuracy, the model cannot generate correct reasoning paths. The generated reasoning path in nearly every sample is of length 1, no matter how many pieces of knowledge are provided. Clearly, the model only learns the format from the prompt and does not know how to decompose questions at all. Note that 1-hop is not considered as multi-hop, we only provide it as a baseline. In the other three settings, the model tries to decompose almost all questions even if it is 1-hop. 

\section{Conclusion}

In this paper, we argue that current LLMs still suffer from many limitations when doing reasoning. First, we show that only depending on the model's internal knowledge is not enough for some reasoning tasks and introducing external knowledge boosts model performance. The absence or presence of distractors has a greater impact on model performance than the proportion of distractors. Moreover, providing knowledge that is inconsistent with the model's internal knowledge confuses the model. Second, LLMs still struggle with problem decomposition and reasoning path generation on non-sequential reasoning tasks even with chain-of-thought prompting (and its variations). Last, LLMs only exhibit limited generalisation ability on the number of reasoning hops. We hope these findings can shed light on the limitations of current models, and encourage improvements in this field.

\section*{Limitations}

All experiments are carried out on text-davinci-002 from OpenAI. In the future, we will try more LLMs. In addition, we only test on small datasets (around 100 samples for each experiment) due to the high cost of the model. We hope that future experiments can be carried out on larger datasets to give more stable results.

\bibliographystyle{unsrt}
\bibliography{custom}

\begin{thebibliography}{10}

\bibitem{devlin2018}
Jacob Devlin, Ming-Wei Chang, Kenton Lee, and Kristina Toutanova.
\newblock Bert: Pre-training of deep bidirectional transformers for language understanding.
\newblock In {\em Proceedings of the 2019 Conference of the North American Chapter of the Association for Computational Linguistics: Human Language Technologies, Volume 1 (Long and Short Papers)}, pages 4171--4186, 2019.

\bibitem{radford2019}
Alec Radford, Jeffrey Wu, Rewon Child, David Luan, Dario Amodei, and Ilya Sutskever.
\newblock Language models are unsupervised multitask learners.
\newblock {\em OpenAI blog}, 1(8):9, 2019.

\bibitem{yang2019}
Zhilin Yang, Zihang Dai, Yiming Yang, Jaime Carbonell, Russ~R Salakhutdinov, and Quoc~V Le.
\newblock Xlnet: Generalized autoregressive pretraining for language understanding.
\newblock In {\em Advances in Neural Information Processing Systems}, 2019.

\bibitem{raffel2020}
Colin Raffel, Noam Shazeer, Adam Roberts, Katherine Lee, Sharan Narang, Michael Matena, Yanqi Zhou, Wei Li, and Peter~J. Liu.
\newblock Exploring the limits of transfer learning with a unified text-to-text transformer.
\newblock {\em Journal of Machine Learning Research}, 21(140):1--67, 2020.

\bibitem{he2023}
Pengcheng He, Jianfeng Gao, and Weizhu Chen.
\newblock Debertav3: Improving deberta using electra-style pre-training with gradient-disentangled embedding sharing.
\newblock {\em arXiv}, arXiv:2111.09543, 2023.

\bibitem{brown2020}
Tom Brown, Benjamin Mann, Nick Ryder, Melanie Subbiah, Jared~D Kaplan, Prafulla Dhariwal, Arvind Neelakantan, Pranav Shyam, Girish Sastry, Amanda Askell, Sandhini Agarwal, Ariel Herbert-Voss, Gretchen Krueger, Tom Henighan, Rewon Child, Aditya Ramesh, Daniel Ziegler, Jeffrey Wu, Clemens Winter, Chris Hesse, Mark Chen, Eric Sigler, Mateusz Litwin, Scott Gray, Benjamin Chess, Jack Clark, Christopher Berner, Sam McCandlish, Alec Radford, Ilya Sutskever, and Dario Amodei.
\newblock Language models are few-shot learners.
\newblock In {\em Advances in Neural Information Processing Systems}, pages 1877--1901, 2020.

\bibitem{chowdhery2023}
Aakanksha Chowdhery, Sharan Narang, Jacob Devlin, Maarten Bosma, Gaurav Mishra, Adam Roberts, Paul Barham, Hyung~Won Chung, Charles Sutton, Sebastian Gehrmann, Parker Schuh, Kensen Shi, Sasha Tsvyashchenko, Joshua Maynez, Abhishek Rao, Parker Barnes, Yi~Tay, Noam Shazeer, Vinodkumar Prabhakaran, Emily Reif, Nan Du, Ben Hutchinson, Reiner Pope, James Bradbury, Jacob Austin, Michael Isard, Guy Gur-Ari, Pengcheng Yin, Toju Duke, Anselm Levskaya, Sanjay Ghemawat, Sunipa Dev, Henryk Michalewski, Xavier Garcia, Vedant Misra, Kevin Robinson, Liam Fedus, Denny Zhou, Daphne Ippolito, David Luan, Hyeontaek Lim, Barret Zoph, Alexander Spiridonov, Ryan Sepassi, David Dohan, Shivani Agrawal, Mark Omernick, Andrew~M. Dai, Thanumalayan~Sankaranarayana Pillai, Marie Pellat, Aitor Lewkowycz, Erica Moreira, Rewon Child, Oleksandr Polozov, Katherine Lee, Zongwei Zhou, Xuezhi Wang, Brennan Saeta, Mark Diaz, Orhan Firat, Michele Catasta, Jason Wei, Kathy Meier-Hellstern, Douglas Eck, Jeff Dean, Slav Petrov, and Noah Fiedel.
\newblock Palm: Scaling language modeling with pathways.
\newblock {\em Journal of Machine Learning Research}, 24(240):1--113, 2023.

\bibitem{openai2024}
OpenAI.
\newblock Gpt-4 technical report.
\newblock {\em arXiv}, arXiv:2303.08774, 2024.

\bibitem{wei2022cot}
Jason Wei, Xuezhi Wang, Dale Schuurmans, Maarten Bosma, brian ichter, Fei Xia, Ed~Chi, Quoc~V Le, and Denny Zhou.
\newblock Chain-of-thought prompting elicits reasoning in large language models.
\newblock In {\em Advances in Neural Information Processing Systems}, pages 24824--24837, 2022.

\bibitem{zhou2023}
Denny Zhou, Nathanael Sch{\"a}rli, Le~Hou, Jason Wei, Nathan Scales, Xuezhi Wang, Dale Schuurmans, Claire Cui, Olivier Bousquet, Quoc~V Le, and Ed~H. Chi.
\newblock Least-to-most prompting enables complex reasoning in large language models.
\newblock In {\em The Eleventh International Conference on Learning Representations}, 2023.

\bibitem{yao2023tot}
Shunyu Yao, Dian Yu, Jeffrey Zhao, Izhak Shafran, Tom Griffiths, Yuan Cao, and Karthik Narasimhan.
\newblock Tree of thoughts: Deliberate problem solving with large language models.
\newblock In {\em Advances in Neural Information Processing Systems}, pages 11809--11822, 2023.

\bibitem{mavi2022}
Vaibhav Mavi, Anubhav Jangra, and Adam Jatowt.
\newblock Multi-hop question answering.
\newblock {\em arXiv}, arXiv:2204.09140, 2022.

\bibitem{staliūnaitė2022}
Ieva Staliūnaitė, Philip~John Gorinski, and Ignacio Iacobacci.
\newblock Relational graph convolutional neural networks for multihop reasoning: A comparative study.
\newblock {\em arXiv}, arXiv:2210.06418, 2022.

\bibitem{zhang2022}
Yanan Zhang, Li~Jin, Xiaoyu Li, and Honqi Wang.
\newblock Edge-aware graph neural network for multi-hop path reasoning over knowledge base.
\newblock {\em Computational Intelligence and Neuroscience}, 2022(1):4734179, 2022.

\bibitem{du2022}
Haowei Du, Quzhe Huang, Chen Zhang, and Dongyan Zhao.
\newblock Knowledge-enhanced iterative instruction generation and reasoning for knowledge base question answering.
\newblock In {\em Natural Language Processing and Chinese Computing: 11th CCF International Conference, NLPCC 2022, Guilin, China, September 24--25, 2022, Proceedings, Part I}, pages 431--444, 2022.

\bibitem{cui2023}
Hai Cui, Tao Peng, Tie Bao, Ridong Han, Jiayu Han, and Lu~Liu.
\newblock Stepwise relation prediction with dynamic reasoning network for multi-hop knowledge graph question answering.
\newblock {\em Applied Intelligence}, 53(10):12340–12354, 2023.

\bibitem{hemmati2023}
Nima Hemmati and Gholamreza Ghassem-Sani.
\newblock Multi-hop question answering using sparse graphs.
\newblock {\em Engineering Applications of Artificial Intelligence}, 126(D):107128, 2023.

\bibitem{xiong2021}
Wenhan Xiong, Xiang Li, Srini Iyer, Jingfei Du, Patrick Lewis, William~Yang Wang, Yashar Mehdad, Scott Yih, Sebastian Riedel, Douwe Kiela, and Barlas Oguz.
\newblock Answering complex open-domain questions with multi-hop dense retrieval.
\newblock In {\em International Conference on Learning Representations}, 2021.

\bibitem{zhao2021}
Chen Zhao, Chenyan Xiong, Jordan Boyd-Graber, and Hal Daum{\'e}~III.
\newblock Multi-step reasoning over unstructured text with beam dense retrieval.
\newblock {\em arXiv}, arXiv:2104.05883, 2021.

\bibitem{wu2022}
Chengmin Wu, Enrui Hu, Ke~Zhan, Lan Luo, Xinyu Zhang, Hao Jiang, Qirui Wang, Zhao Cao, Fan Yu, and Lei Chen.
\newblock Triple-fact retriever: An explainable reasoning retrieval model for multi-hop qa problem.
\newblock {\em 2022 IEEE 38th International Conference on Data Engineering (ICDE)}, 53(10):1206--1218, 2022.

\bibitem{xu2021}
Yichong Xu, Chenguang Zhu, Shuohang Wang, Siqi Sun, Hao Cheng, Xiaodong Liu, Jianfeng Gao, Pengcheng He, Michael Zeng, and Xuedong Huang.
\newblock Human parity on commonsenseqa: Augmenting self-attention with external attention.
\newblock {\em arXiv}, arXiv:2112.03254, 2021.

\bibitem{wang2021}
Sinong Wang, Han Fang, Madian Khabsa, Hanzi Mao, and Haog Ma.
\newblock Entailment as few-shot learner.
\newblock {\em arXiv}, arXiv:2104.14690, 2021.

\bibitem{min2019}
Sewon Min, Victor Zhong, Luke Zettlemoyer, and Hannaneh Hajishirzi.
\newblock Multi-hop reading comprehension through question decomposition and rescoring.
\newblock In {\em Proceedings of the 57th Annual Meeting of the Association for Computational Linguistics}, pages 6097--6109, 2019.

\bibitem{press2023}
Ofir Press, Muru Zhang, Sewon Min, Ludwig Schmidt, Noah Smith, and Mike Lewis.
\newblock Measuring and narrowing the compositionality gap in language models.
\newblock In {\em Findings of the Association for Computational Linguistics: EMNLP 2023}, pages 5687--5711, 2023.

\bibitem{zhang2024}
Kun Zhang, Jiali Zeng, Fandong Meng, Yuanzhuo Wang, Shiqi Sun, Long Bai, Huawei Shen, and Jie Zhou.
\newblock Tree-of-reasoning question decomposition for complex question answering with large language models.
\newblock In {\em Proceedings of the AAAI Conference on Artificial Intelligence}, pages 19560--19568, 2024.

\bibitem{touvron2023}
Hugo Touvron, Louis Martin, Kevin Stone, Peter Albert, Amjad Almahairi, Yasmine Babaei, Nikolay Bashlykov, Soumya Batra, Prajjwal Bhargava, Shruti Bhosale, Dan Bikel, Lukas Blecher, Cristian~Canton Ferrer, Moya Chen, Guillem Cucurull, David Esiobu, Jude Fernandes, Jeremy Fu, Wenyin Fu, Brian Fuller, Cynthia Gao, Vedanuj Goswami, Naman Goyal, Anthony Hartshorn, Saghar Hosseini, Rui Hou, Hakan Inan, Marcin Kardas, Viktor Kerkez, Madian Khabsa, Isabel Kloumann, Artem Korenev, Punit~Singh Koura, Marie-Anne Lachaux, Thibaut Lavril, Jenya Lee, Diana Liskovich, Yinghai Lu, Yuning Mao, Xavier Martinet, Todor Mihaylov, Pushkar Mishra, Igor Molybog, Yixin Nie, Andrew Poulton, Jeremy Reizenstein, Rashi Rungta, Kalyan Saladi, Alan Schelten, Ruan Silva, Eric~Michael Smith, Ranjan Subramanian, Xiaoqing~Ellen Tan, Binh Tang, Ross Taylor, Adina Williams, Jian~Xiang Kuan, Puxin Xu, Zheng Yan, Iliyan Zarov, Yuchen Zhang, Angela Fan, Melanie Kambadur, Sharan Narang, Aurelien Rodriguez, Robert Stojnic, Sergey Edunov, and Thomas
  Scialom.
\newblock Llama 2: Open foundation and fine-tuned chat models.
\newblock {\em arXiv}, arXiv:2307.09288, 2023.

\bibitem{bang2023}
Yejin Bang, Samuel Cahyawijaya, Nayeon Lee, Wenliang Dai, Dan Su, Bryan Wilie, Holy Lovenia, Ziwei Ji, Tiezheng Yu, Willy Chung, Quyet~V. Do, Yan Xu, and Pascale Fung.
\newblock A multitask, multilingual, multimodal evaluation of chatgpt on reasoning, hallucination, and interactivity.
\newblock {\em arXiv}, arXiv:2302.04023, 2023.

\bibitem{wei2022}
Jason Wei, Maarten Bosma, Vincent Zhao, Kelvin Guu, Adams~Wei Yu, Brian Lester, Nan Du, Andrew~M. Dai, and Quoc~V Le.
\newblock Finetuned language models are zero-shot learners.
\newblock In {\em International Conference on Learning Representations}, 2022.

\bibitem{sanh2022}
Victor Sanh, Albert Webson, Colin Raffel, Stephen Bach, Lintang Sutawika, Zaid Alyafeai, Antoine Chaffin, Arnaud Stiegler, Arun Raja, Manan Dey, M~Saiful Bari, Canwen Xu, Urmish Thakker, Shanya~Sharma Sharma, Eliza Szczechla, Taewoon Kim, Gunjan Chhablani, Nihal Nayak, Debajyoti Datta, Jonathan Chang, Mike Tian-Jian Jiang, Han Wang, Matteo Manica, Sheng Shen, Zheng~Xin Yong, Harshit Pandey, Rachel Bawden, Thomas Wang, Trishala Neeraj, Jos Rozen, Abheesht Sharma, Andrea Santilli, Thibault Fevry, Jason~Alan Fries, Ryan Teehan, Teven~Le Scao, Stella Biderman, Leo Gao, Thomas Wolf, and Alexander~M Rush.
\newblock Multitask prompted training enables zero-shot task generalization.
\newblock In {\em International Conference on Learning Representations}, 2022.

\bibitem{wang2023}
Xuezhi Wang, Jason Wei, Dale Schuurmans, Quoc~V Le, Ed~H. Chi, Sharan Narang, Aakanksha Chowdhery, and Denny Zhou.
\newblock Self-consistency improves chain of thought reasoning in language models.
\newblock In {\em The Eleventh International Conference on Learning Representations}, 2023.

\bibitem{kojima2022}
Takeshi Kojima, Shixiang~(Shane) Gu, Machel Reid, Yutaka Matsuo, and Yusuke Iwasawa.
\newblock Large language models are zero-shot reasoners.
\newblock In {\em Advances in Neural Information Processing Systems}, pages 22199--22213, 2022.

\bibitem{creswell2023}
Antonia Creswell, Murray Shanahan, and Irina Higgins.
\newblock Selection-inference: Exploiting large language models for interpretable logical reasoning.
\newblock In {\em The Eleventh International Conference on Learning Representations}, 2023.

\bibitem{yao2023react}
Shunyu Yao, Jeffrey Zhao, Dian Yu, Nan Du, Izhak Shafran, Karthik~R Narasimhan, and Yuan Cao.
\newblock React: Synergizing reasoning and acting in language models.
\newblock In {\em The Eleventh International Conference on Learning Representations}, 2023.

\bibitem{yang2018}
Zhilin Yang, Peng Qi, Saizheng Zhang, Yoshua Bengio, William Cohen, Ruslan Salakhutdinov, and Christopher~D. Manning.
\newblock Hotpotqa: A dataset for diverse, explainable multi-hop question answering.
\newblock In {\em Proceedings of the 2018 Conference on Empirical Methods in Natural Language Processing}, pages 2369--2380, 2018.

\bibitem{dalvi2021}
Bhavana Dalvi, Peter Jansen, Oyvind Tafjord, Zhengnan Xie, Hannah Smith, Leighanna Pipatanangkura, and Peter Clark.
\newblock Explaining answers with entailment trees.
\newblock In {\em Proceedings of the 2021 Conference on Empirical Methods in Natural Language Processing}, pages 7358--7370, 2021.

\bibitem{khot2020}
Tushar Khot, Peter Clark, Michal Guerquin, Peter Jansen, and Ashish Sabharwal.
\newblock Qasc: A dataset for question answering via sentence composition.
\newblock In {\em Proceedings of the AAAI Conference on Artificial Intelligence}, pages 8082--8090, 2020.

\bibitem{weston2015}
Jason Weston, Antoine Bordes, Sumit Chopra, Alexander~M Rush, Bart Van~Merri{\"e}nboer, Armand Joulin, and Tomas Mikolov.
\newblock Towards ai-complete question answering: A set of prerequisite toy tasks.
\newblock {\em arXiv}, arXiv:1502.05698, 2015.

\bibitem{van2015}
Nicole Van~Hoeck, Patrick~D Watson, and Aron~K Barbey.
\newblock Cognitive neuroscience of human counterfactual reasoning.
\newblock {\em Frontiers in human neuroscience}, 9:420, 2015.

\bibitem{yin2023}
Zhangyue Yin, Qiushi Sun, Qipeng Guo, Jiawen Wu, Xipeng Qiu, and Xuanjing Huang.
\newblock Do large language models know what they don{'}t know?
\newblock In {\em Findings of the Association for Computational Linguistics: ACL 2023}, pages 8653--8665, 2023.

\bibitem{jansen2017}
Peter Jansen.
\newblock A study of automatically acquiring explanatory inference patterns from corpora of explanations: Lessons from elementary science exams.
\newblock In {\em AKBC @ NIPS}, 2017.

\bibitem{min2022}
Sewon Min, Xinxi Lyu, Ari Holtzman, Mikel Artetxe, Mike Lewis, Hannaneh Hajishirzi, and Luke Zettlemoyer.
\newblock Rethinking the role of demonstrations: What makes in-context learning work?
\newblock In {\em Proceedings of the 2022 Conference on Empirical Methods in Natural Language Processing}, pages 11048--11064, 2022.

\end{thebibliography}

\appendix

\section{Appendix}

\subsection{Prompts for HotpotQA}

\textbf{No-context}

~\\
\noindent
\textbf{Q:}
According to the 2010 census, what was the population of the city in which Boyle's Thirty Acres was located?

\noindent
\textbf{A:}
Boyles' Thirty Acres was in Jersey City, New Jersey. And the population of Jersey City, New Jersey was 247,597 in 2010.
Therefore, the answer is 247,597.

~\\
\textbf{Q:}
What basketball player elected to the National Collegiate Basketball Hall of Fame released music through Rendezvous Music?

\noindent
\textbf{A:}
Kirk Whalum, Jonathan Butler, Kyle Eastwood and Wayman Tisdale released music through Rendezvous Music. And Wayman Tisdale elected to the National Collegiate Basketball Hall of Fame.
Therefore, the answer is Wayman Tisdale.

~\\
\textbf{Q:}
Who commanded the Anglo-Canadian 21st Army Group and was nicknamed `Monty'?

\noindent
\textbf{A:}
Bernard Montgomery commanded the Anglo-Canadian 21st Army Group. And Bernard Montgomery was nicknamed `Monty'.
Therefore, the answer is Bernard Montgomery.

~\\
\textbf{Q:}
Who was one of the star in the South Korean horror romantic comedy film of Hwang In-ho?

\noindent
\textbf{A:}
The horror romantic comedy film of Hwang In-ho is `Spellbound' (2011). And Son Ye-jin starred in this film.
Therefore, the answer is Son Ye-jin.

~\\
\noindent
\textbf{100\%-gold}

~\\
\noindent
\textbf{C:} 
Content 1: Boyle's Thirty Acres was a large wooden bowl arena in Jersey City, New Jersey.

\noindent
Content 2: As of 2016, the Census Bureau's Population Estimates Program calculated that Jersey City's population was 264,152, with the largest population increase of any municipality in New Jersey since 2010, an increase of about 6.7\% from the 2010 United States Census, when the city's population was at 247,597, ranking the city the 77th-largest in the nation.

\noindent
\textbf{Q:} According to the 2010 census, what was the population of the city in which Boyle's Thirty Acres was located?

\noindent
\textbf{A:} From Content 1 we know Boyles' Thirty Acres was in Jersey City, New Jersey. From Content 2 we know the population of Jersey City, New Jersey was 247,597 in 2010. Therefore, the selected supporting paragraphs are 1 and 2. And the answer is 247,597.

~\\
\textbf{C:}
Content 1: Rendezvous Music (formerly known as Rendezvous Entertainment) is a record label founded by multiple-Grammy Award nominee and RIAA Certified Gold recording artist Dave Koz, radio entrepreneur Frank Cody and music business veteran Hyman Katz in 2002. Since its acquisition in August 2008, Rendezvous is part of the Mack Avenue Records label group and continues to release new music by Kirk Whalum, Jonathan Butler, Kyle Eastwood as well as previously unreleased music by Wayman Tisdale.

\noindent
Content 2: Wayman Lawrence Tisdale (June 9, 1964 – May 15, 2009) was an American professional basketball player in the NBA and a smooth jazz bass guitarist. A three-time All American at the University of Oklahoma, he was elected to the National Collegiate Basketball Hall of Fame in 2009.

\noindent
\textbf{Q:} What basketball player elected to the National Collegiate Basketball Hall of Fame released music through Rendezvous Music?

\noindent
\textbf{A:} From Content 1 we know Kirk Whalum, Jonathan Butler, Kyle Eastwood and Wayman Tisdale released music through Rendezvous Music. From Content 2 we know was Wayman Tisdale elected to the National Collegiate Basketball Hall of Fame. Therefore, the selected supporting paragraphs are 1 and 2. And the answer is Wayman Tisdale.

~\\
\textbf{C:}
Content 1: Derek Knee (22 October 1922 – 18 March 2014) was a British Army intelligence officer during World War II who was the interpreter and translator for Field Marshal Sir Bernard Montgomery, commander of the Anglo-Canadian 21st Army Group, at the German surrender at Lüneburg Heath in Germany on 3 May 1945.

\noindent
Content 2: Field Marshal Bernard Law Montgomery, 1st Viscount Montgomery of Alamein, (17 November 1887 – 24 March 1976), nicknamed `Monty' and the `Spartan General', was a senior British Army officer who fought in both the First World War and the Second World War.

\noindent
\textbf{Q:} Who commanded the Anglo-Canadian 21st Army Group and was nicknamed `Monty'?

\noindent
\textbf{A:} From Content 1 we know Bernard Montgomery commanded the Anglo-Canadian 21st Army Group. From Content 2 we know Bernard Montgomery was nicknamed `Monty'. Therefore, the selected supporting paragraphs are 1 and 2. And the answer is Bernard Montgomery.

~\\
\textbf{C:}
Content 1: Hwang In-ho is a South Korean film director and screenwriter. He wrote and directed the horror-romantic comedy `Spellbound' (2011) and the crime thriller `Monster' (2014).

\noindent
Content 2: Spellbound (`Chilling Romance') is a 2011 South Korean horror romantic comedy film, starring Son Ye-jin.

\noindent
\textbf{Q:} Who was one of the star in the South Korean horror romantic comedy film of Hwang In-ho?

\noindent
\textbf{A:} From Content 1 we know the horror romantic comedy film of Hwang In-ho is `Spellbound' (2011). From Content 2 we know Son Ye-jin starred in this film. Therefore, the selected supporting paragraphs are 1 and 2. And the answer is Son Ye-jin.

~\\
\textbf{50\%-gold}

~\\
\textbf{C:}
Content 1: Largely residential in character, Murrieta is typically characterized as a commuter town, with many of its residents commuting to jobs in San Diego County, Orange County, Los Angeles County, Temecula, and Camp Pendleton.

\noindent
Content 2: Boyle's Thirty Acres was a large wooden bowl arena in Jersey City, New Jersey.

\noindent
Content 3: As of 2016, the Census Bureau's Population Estimates Program calculated that Jersey City's population was 264,152, with the largest population increase of any municipality in New Jersey since 2010, an increase of about 6.7\% from the 2010 United States Census, when the city's population was at 247,597, ranking the city the 77th-largest in the nation.

\noindent
Content 4: It has 4,741,948 of population (2010 census), shares 7.22\% of the province.

\noindent
\textbf{Q:} According to the 2010 census, what was the population of the city in which Boyle's Thirty Acres was located?

\noindent
\textbf{A:} From Content 2 we know Boyles' Thirty Acres was in Jersey City, New Jersey. From Content 3 we know the population of Jersey City, New Jersey was 247,597 in 2010. Therefore, the selected supporting paragraphs are 2 and 3. And the answer is 247,597.

~\\
\textbf{C:}
Content 1: Rendezvous Music (formerly known as Rendezvous Entertainment) is a record label founded by multiple-Grammy Award nominee and RIAA Certified Gold recording artist Dave Koz, radio entrepreneur Frank Cody and music business veteran Hyman Katz in 2002. Since its acquisition in August 2008, Rendezvous is part of the Mack Avenue Records label group and continues to release new music by Kirk Whalum, Jonathan Butler, Kyle Eastwood as well as previously unreleased music by Wayman Tisdale.

\noindent
Content 2: He served as the head men's basketball coach at the University of Houston from 1956 to 1986. His 1980s teams, nicknamed Phi Slama Jama for their slam dunks, were runners-up for the national championship in back-to-back seasons in 1983 and 1984.

\noindent
Content 3: Wayman Lawrence Tisdale (June 9, 1964 – May 15, 2009) was an American professional basketball player in the NBA and a smooth jazz bass guitarist. A three-time All American at the University of Oklahoma, he was elected to the National Collegiate Basketball Hall of Fame in 2009.

\noindent
Content 4: He is known as the Godfather of Florida basketball. He coached from 1957 to 1993. He was the coach of the Stetson Hatters and helped in their transition from the NAIA to NCAA Division I.

\noindent
\textbf{Q:} What basketball player elected to the National Collegiate Basketball Hall of Fame released music through Rendezvous Music?

\noindent
\textbf{A:} From Content 1 we know Kirk Whalum, Jonathan Butler, Kyle Eastwood and Wayman Tisdale released music through Rendezvous Music. From Content 3 we know was Wayman Tisdale elected to the National Collegiate Basketball Hall of Fame. Therefore, the selected supporting paragraphs are 1 and 3. And the answer is Wayman Tisdale.

~\\
\textbf{C:}
Content 1: Its history can be traced to the 307th Division, 100th Corps, 1st Army Group of Republic of China Army defected in August 1948.

\noindent
Content 2: Broken up in mid-August, it was one of two divisions of the army group that was disbanded due to a very severe shortage of manpower in the British Army.

\noindent
Content 3: Derek Knee (22 October 1922 – 18 March 2014) was a British Army intelligence officer during World War II who was the interpreter and translator for Field Marshal Sir Bernard Montgomery, commander of the Anglo-Canadian 21st Army Group, at the German surrender at Lüneburg Heath in Germany on 3 May 1945.

\noindent
Content 4: Field Marshal Bernard Law Montgomery, 1st Viscount Montgomery of Alamein, (17 November 1887 – 24 March 1976), nicknamed `Monty' and the `Spartan General', was a senior British Army officer who fought in both the First World War and the Second World War.

\noindent
\textbf{Q:} Who commanded the Anglo-Canadian 21st Army Group and was nicknamed `Monty'?

\noindent
\textbf{A:} From Content 3 we know Bernard Montgomery commanded the Anglo-Canadian 21st Army Group. From Content 4 we know Bernard Montgomery was nicknamed `Monty'. Therefore, the selected supporting paragraphs are 3 and 4. And the answer is Bernard Montgomery.

~\\
\textbf{C:}
Content 1: Hwang In-ho is a South Korean film director and screenwriter. He wrote and directed the horror-romantic comedy `Spellbound' (2011) and the crime thriller `Monster' (2014).

\noindent
Content 2: Hwang Bo-ra (born October 2, 1983) is a South Korean actress.

\noindent
Content 3: The series uses an all-girls high school as the backdrop for each of its films and doesn't share a continuing plot.

\noindent
Content 4: Spellbound (lit. `Chilling Romance') is a 2011 South Korean horror romantic comedy film, starring Son Ye-jin and Lee Min-ki.

\noindent
\textbf{Q:} Who was one of the star in the South Korean horror romantic comedy film of Hwang In-ho?

\noindent
\textbf{A:} From Content 1 we know the horror romantic comedy film of Hwang In-ho is `Spellbound' (2011). From Content 4 we know Son Ye-jin starred in this film. Therefore, the selected supporting paragraphs are 1 and 4. And the answer is Son Ye-jin.

~\\
\textbf{20\%-gold}

~\\
\textbf{C:}
Content 1: Largely residential in character, Murrieta is typically characterized as a commuter town, with many of its residents commuting to jobs in San Diego County, Orange County, Los Angeles County, Temecula, and Camp Pendleton.

\noindent
Content 2: Dune Acres is a town in Westchester Township, Porter County, in the U.S. state of Indiana. The population was 182 at the 2010 census.

\noindent
Content 3: Boyle County is a county located in the U.S. state of Kentucky. As of the 2010 census, the population was 28,432. Its county seat is Danville.

\noindent
Content 4: Mankato is a city in Blue Earth, Nicollet, and Le Sueur counties in the state of Minnesota. The population was 41,044 according to 2015 US census estimates, making it the fifth largest city in Minnesota outside of the Minneapolis-Saint Paul metropolitan area.

\noindent
Content 5: Boyle's Thirty Acres was a large wooden bowl arena in Jersey City, New Jersey.

\noindent
Content 6: As of 2016, the Census Bureau's Population Estimates Program calculated that Jersey City's population was 264,152, with the largest population increase of any municipality in New Jersey since 2010, an increase of about 6.7\% from the 2010 United States Census, when the city's population was at 247,597, ranking the city the 77th-largest in the nation.

\noindent
Content 7: The city of Union is the county seat of Union County, South Carolina, United States. The population was 8,393 at the 2010 census.

\noindent
Content 8: It has 4,741,948 of population (2010 census), shares 7.22\% of the province.

\noindent
Content 9: Altoona is a city in Blair County, Pennsylvania, United States. The population was 46,320 at the time of the 2010 Census, making it the eleventh most populous city in Pennsylvania.

\noindent
Content 10: Ubinsky District is an administrative and municipal district (raion), one of the thirty in Novosibirsk Oblast, Russia. Population: 16,297 (2010 Census) The population of Ubinskoye accounts for 35.8\% of the district's total population.

\noindent
\textbf{Q:} According to the 2010 census, what was the population of the city in which Boyle's Thirty Acres was located?

\noindent
\textbf{A:} From Content 5 we know Boyles' Thirty Acres was in Jersey City, New Jersey. From Content 6 we know the population of Jersey City, New Jersey was 247,597 in 2010. Therefore, the selected supporting paragraphs are 5 and 6. And the answer is 247,597.

~\\
\textbf{C:}
Content 1: Rendezvous Music (formerly known as Rendezvous Entertainment) is a record label founded by multiple-Grammy Award nominee and RIAA Certified Gold recording artist Dave Koz, radio entrepreneur Frank Cody and music business veteran Hyman Katz in 2002. Since its acquisition in August 2008, Rendezvous is part of the Mack Avenue Records label group and continues to release new music by Kirk Whalum, Jonathan Butler, Kyle Eastwood as well as previously unreleased music by Wayman Tisdale.

\noindent
Content 2: He served as the head men's basketball coach at the University of Houston from 1956 to 1986. His 1980s teams, nicknamed Phi Slama Jama for their slam dunks, were runners-up for the national championship in back-to-back seasons in 1983 and 1984.

\noindent
Content 3: Hickey also served as the head football coach at Creighton in 1934, tallying a mark of 2-7. After retiring from coaching, Hickey managed the American Automobile Association headquarters in Terre Haute, Indiana.

\noindent
Content 4: He is known as the Godfather of Florida basketball. He coached from 1957 to 1993. He was the coach of the Stetson Hatters and helped in their transition from the NAIA to NCAA Division I.

\noindent
Content 5: Gallatin played nine seasons for the New York Knicks in the NBA from 1948 to 1957, as well as one season with the Detroit Pistons in the 1957-58 season. Gallatin led the NBA in rebounding and was named to the All-NBA First Team in 1954.

\noindent
Content 6: Nolan Richardson (born December 27, 1941) is a former American basketball head coach best known for his tenure at the University of Arkansas, where he won the 1994 NCAA Men's Division I Basketball Tournament. During his 22 seasons of coaching in NCAA Division I, Richardson made a post-season tournament appearance 20 times.

\noindent
Content 7: Tom Jernstedt is an American basketball administrator, working for the NCAA from 1972 until 2010. He was enshrined into the National Collegiate Basketball Hall of Fame as a contributor in 2010 and the Naismith Memorial Basketball Hall of Fame in 2017.

\noindent
Content 8: He was a major contributor to the development of modern basketball and coached on both the college and professional levels during his career. He has been enshrined twice in the Naismith Memorial Basketball Hall of Fame, and also inducted into the National Collegiate Basketball Hall of Fame.

\noindent
Content 9: Wayman Lawrence Tisdale (June 9, 1964 – May 15, 2009) was an American professional basketball player in the NBA and a smooth jazz bass guitarist. A three-time All American at the University of Oklahoma, he was elected to the National Collegiate Basketball Hall of Fame in 2009.

\noindent
Content 10: The museum is an integral portion of the College Basketball Experience created by the National Association of Basketball Coaches (NABC), located at the Sprint Center. The hall is meant as a complement to the Naismith Memorial Basketball Hall of Fame, with a focus strictly on those who have contributed greatly to college basketball.

\noindent
\textbf{Q:} What basketball player elected to the National Collegiate Basketball Hall of Fame released music through Rendezvous Music?

\noindent
\textbf{A:} From Content 1 we know Kirk Whalum, Jonathan Butler, Kyle Eastwood and Wayman Tisdale released music through Rendezvous Music. From Content 9 we know was Wayman Tisdale elected to the National Collegiate Basketball Hall of Fame. Therefore, the selected supporting paragraphs are 1 and 9. And the answer is Wayman Tisdale.

~\\
\textbf{C:}
Content 1: The 218th Division was created in November 1949 under `the Regulation of the Redesignations of All Organizations and Units of the Army', issued by Central Military Commission on November 1, 1948, basing on the 3rd and 7th Division, 3rd Corps, 1st Army Group of the People's Liberation Army of the Nationalist Party of China.

\noindent
Content 2: Its history can be traced to the 307th Division, 100th Corps, 1st Army Group of Republic of China Army defected in August 1948.

\noindent
Content 3: He was Deputy Commander of the lines of communication of the 21st Army Group from May to November 1944, and then commanded the lines of communication in South East Asia Command (SEAC).

\noindent
Content 4: Broken up in mid-August, it was one of two divisions of the army group that was disbanded due to a very severe shortage of manpower in the British Army.

\noindent
Content 5: Derek Knee (22 October 1922 – 18 March 2014) was a British Army intelligence officer during World War II who was the interpreter and translator for Field Marshal Sir Bernard Montgomery, commander of the Anglo-Canadian 21st Army Group, at the German surrender at Lüneburg Heath in Germany on 3 May 1945.

\noindent
Content 6: `Veritable' (originally called `Valediction') had been planned for execution in early January, 1945 when the ground had been frozen and thus more advantageous to the Allies.

\noindent
Content 7: The 21st Army Group was a World War II British headquarters formation, in command of two field armies and other supporting units, consisting primarily of the British Second Army and the First Canadian Army.

\noindent
Content 8: Its history can be traced to the 63rd Division, 14th Corps, 1st Army Group of Republic of China Army defected in August 1948.

\noindent
Content 9: Field Marshal Bernard Law Montgomery, 1st Viscount Montgomery of Alamein, (17 November 1887 – 24 March 1976), nicknamed `Monty' and the `Spartan General', was a senior British Army officer who fought in both the First World War and the Second World War.

\noindent
Content 10: The 214th Division was created in November 1949 under `the Regulation of the Redesignations of All Organizations and Units of the Army', issued by Central Military Commission on November 1, 1948, basing on the 1st Division, 1st Corps, 1st Army Group of the People's Liberation Army of the Nationalist Party of China.

\noindent
\textbf{Q:} Who commanded the Anglo-Canadian 21st Army Group and was nicknamed `Monty'?

\noindent
\textbf{A:} From Content 5 we know Bernard Montgomery commanded the Anglo-Canadian 21st Army Group. From Content 9 we know Bernard Montgomery was nicknamed `Monty'. Therefore, the selected supporting paragraphs are 5 and 9. And the answer is Bernard Montgomery.

~\\
\noindent
\textbf{C:}
Content 1: Hwang has since played leading roles in the indie melodrama `Lovers Vanished' (2010), the TV sitcom `I Need a Fairy' (also known as `Sent from Heaven', 2012), and the romantic comedy `Virgin Theory: 7 Steps to Get On the Top' (2014).

\noindent
Content 2: The film tells the story of a Catholic priest—who is in love with his friend’s wife—turning into a vampire through a failed medical experiment.

\noindent
Content 3: It was produced by JK Film and distributed by CJ Entertainment, and released on January 18, 2012.

\noindent
Content 4: Whispering Corridors is a 1998 South Korean horror film.

\noindent
Content 5: Hwang In-ho is a South Korean film director and screenwriter. He wrote and directed the horror-romantic comedy `Spellbound' (2011) and the crime thriller `Monster' (2014).

\noindent
Content 6: Hwang Bo-ra (born October 2, 1983) is a South Korean actress.

\noindent
Content 7: The series uses an all-girls high school as the backdrop for each of its films and doesn't share a continuing plot.

\noindent
Content 8: Spellbound (`Chilling Romance') is a 2011 South Korean horror romantic comedy film, starring Son Ye-jin and Lee Min-ki.

\noindent
Content 9: `The Ghost' is a 2004 South Korean horror film.

\noindent
Content 10: Monster is a 2014 South Korean thriller film written and directed by Hwang In-ho, starring Lee Min-ki and Kim Go-eun.

\noindent
\textbf{Q:} Who was one of the star in the South Korean horror romantic comedy film of Hwang In-ho?

\noindent
\textbf{A:} From Content 5 we know the horror romantic comedy film of Hwang In-ho is `Spellbound' (2011). From Content 8 we know Son Ye-jin starred in this film. Therefore, the selected supporting paragraphs are 5 and 8. And the answer is Son Ye-jin.

\subsection{Prompts for EntailmentBank}

\textbf{No-context}

~\\
\textbf{Q:} How is the structure of an atom best described?

\noindent
\textbf{A:} Protons and neutrons together will be positive in charge. The nucleus of an atom is positive in charge. An atom has a positve charged core. The core of an atom is surrounded by almost empty space. An atom has a positive core surrounded by mostly empty space. Therefore, the answer is a positive core surrounded by mostly empty space.

~\\
\textbf{Q:} A beaker containing 50 milliliters of ice is placed on a windowsill. After several hours, the ice melts. What property of the ice did not change when it melted?

\noindent
\textbf{A:} The ice undergoes a phase change. The mass of the ice will not change. Therefore, the answer is the mass.

~\\
\textbf{Q:} Natural processes cause rapid and slow changes to Earth's surface. Which event causes changes to Earth's surface most rapidly?

\noindent
\textbf{A:} An earthquake usually occurs in a short amount of time. An earthquake can change earth's surface by shaking the ground. An earthquake can change earth's surface in a short amount of time. An earthquake changes earth 's surface quickly. Therefore, the answer is an earthquake shaking the ground.

~\\
\textbf{Q:} What makes solar energy different from most other energy that people use?

\noindent
\textbf{A:} The supply of solar energy will not change over a long period of time. To a human, a time period of billions of years is considered a very long time. The supply of solar energy will not change over billions of years. Therefore, the answer is the supply of solar energy will not change for billions of years.

~\\
\textbf{100\%-gold}

~\\
\textbf{C:} sent1: the nucleus of an atom is made of protons and neutrons. sent2: the nucleus of an atom is surrounded by almost empty space. sent3: the nucleus is the core of an atom. sent4: a neutron has a neutral electric charge. sent5: a proton has a positive 1 electric charge. 

\noindent
\textbf{Q:} How is the structure of an atom best described?

\noindent
\textbf{A:} The reasoning path is: sent4 \& sent5 -> int1: protons and neutrons together will be positive in charge; int1 \& sent1 -> int2: the nucleus of an atom is positive in charge; int2 \& sent3 -> int3: an atom has a positve charged core; sent2 \& sent3 -> int4: the core of an atom is surrounded by almost empty space; int3 \& int4 -> an atom has a positive core surrounded by mostly empty space. Therefore, the answer is a positive core surrounded by mostly empty space.

~\\
\textbf{C:} sent1: melting is a kind of phase change. sent2: the ice melts. sent3: phase changes do not change mass.

\noindent
\textbf{Q:} A beaker containing 50 milliliters of ice is placed on a windowsill. After several hours, the ice melts. What property of the ice did not change when it melted?

\noindent
\textbf{A:} The reasoning path is: sent1 \& sent2 -> int1: the ice undergoes a phase change; int1 \& sent3 -> the mass of the ice will not change. Therefore, the answer is the mass.

~\\
\textbf{C:} sent1: shaking the ground of the earth can change earth's surface. sent2: an earthquake causes the ground to shake. sent3: in a short amount of time is similar to quickly. sent4: an earthquake usually occurs over a period of 10 to 30 seconds. sent5: 10 to 30 seconds is a short amount of time. 

\noindent
\textbf{Q:} Natural processes cause rapid and slow changes to Earth's surface. Which event causes changes to Earth's surface most rapidly?

\noindent
\textbf{A:} The reasoning path is: sent4 \& sent5 -> int1: an earthquake usually occurs in a short amount of time; sent2 \& sent1 -> int2: an earthquake can change earth's surface by shaking the ground; int1 \& int2 -> int3: an earthquake can change earth's surface in a short amount of time; int3 \& sent3 -> an earthquake changes earth 's surface quickly. Therefore, the answer is an earthquake shaking the ground.

~\\
\textbf{C:} sent1: billions of years is a longer period of time than 10 years. sent2: solar energy is a renewable / inexhaustible resource. sent3: the supply of an inexhaustible resource does not change over a long period of time. sent4: to a human, a time period of 10 years or more years is considered very long. 

\noindent
\textbf{Q:} What makes solar energy different from most other energy that people use?

\noindent
\textbf{A:} The reasoning path is: sent2 \& sent3 -> int1: the supply of solar energy will not change over a long period of time; sent4 \& sent1 -> int2: to a human, a time period of billions of years is considered a very long time; int1 \& int2 -> the supply of solar energy will not change over billions of years. Therefore, the answer is the supply of solar energy will not change for billions of years.

~\\
\textbf{50\%-gold}

~\\
\textbf{C:} sent1: electrons are located at the outside of the nucleus. sent2: best means most positive. sent3: the nucleus of an atom is made of protons and neutrons. sent4: the nucleus of an atom is surrounded by almost empty space. sent5: the nucleus is the core of an atom. sent6: a neutron has a neutral electric charge. sent7: nuclear fusion happens in the core of a star. sent8: a proton has a positive 1 electric charge. sent9: the properties of something can be used to identify / used to describe that something. sent10: electron configuration is a property of elements.

\noindent
\textbf{Q:} How is the structure of an atom best described?

\noindent
\textbf{A:} The reasoning path is: sent6 \& sent8 -> int1: protons and neutrons together will be positive in charge; int1 \& sent3 -> int2: the nucleus of an atom is positive in charge; int2 \& sent5 -> int3: an atom has a positve charged core; sent4 \& sent5 -> int4: the core of an atom is surrounded by almost empty space; int3 \& int4 -> an atom has a positive core surrounded by mostly empty space. Therefore, the answer is a positive core surrounded by mostly empty space.

~\\
\textbf{C:} sent1: rocks in liquid state is similar to melted rock. sent2: melting is a kind of phase change. sent3: the ice melts. sent4: hours are a kind of unit for measuring time. sent5: phase changes do not change mass. sent6: melting point is a property of a substance / material. 

\noindent
\textbf{Q:} A beaker containing 50 milliliters of ice is placed on a windowsill. After several hours, the ice melts. What property of the ice did not change when it melted?

\noindent
\textbf{A:} The reasoning path is: sent2 \& sent3 -> int1: the ice undergoes a phase change; int1 \& sent5 -> the mass of the ice will not change. Therefore, the answer is the mass.

~\\
\textbf{C:} sent1: tectonic plates being pushed together causes earthquakes. sent2: an earthquake wave is a kind of wave. sent3: motion / movement means moving / to move. sent4: shaking the ground of the earth can change earth's surface. sent5: to cause means to result in. sent6: an earthquake causes the ground to shake. sent7: in a short amount of time is similar to quickly. sent8: an earthquake usually occurs over a period of 10 to 30 seconds. sent9: the ground means earth's surface. sent10: 10 to 30 seconds is a short amount of time.

\noindent
\textbf{Q:} Natural processes cause rapid and slow changes to Earth's surface. Which event causes changes to Earth's surface most rapidly?

\noindent
\textbf{A:} The reasoning path is: sent8 \& sent10 -> int1: an earthquake usually occurs in a short amount of time; sent6 \& sent4 -> int2: an earthquake can change earth's surface by shaking the ground; int1 \& int2 -> int3: an earthquake can change earth's surface in a short amount of time; int3 \& sent7 -> an earthquake changes earth 's surface quickly. Therefore, the answer is the behavior is an earthquake shaking the ground.

~\\
\textbf{C:} sent1: the sun is the source of energy for life on earth. sent2: solar power means solar energy. sent3: billions of years is a longer period of time than 10 years. sent4: solar energy is a renewable / inexhaustible resource. sent5: photosynthesis means producers / green plants convert from carbon dioxide and water and solar energy into carbohydrates and food and oxygen for themselves. sent6: lifetime is similar to life cycle. sent7: the supply of an inexhaustible resource does not change over a long period of time. sent8: to a human, a time period of 10 years or more years is considered very long. 

\noindent
\textbf{Q:} What makes solar energy different from most other energy that people use?

\noindent
\textbf{A:} The reasoning path is: sent4 \& sent7 -> int1: the supply of solar energy will not change over a long period of time; sent8 \& sent3 -> int2: to a human, a time period of billions of years is considered a very long time; int1 \& int2 -> the supply of solar energy will not change over billions of years. Therefore, the answer is the supply of solar energy will not change for billions of years.

~\\
\textbf{20\%-gold}

~\\
\textbf{C:} sent1: nucleus is synonymous with core / center. sent2: electrons are located at the outside of the nucleus. sent3: state of matter is a property of matter and includes ordered values of solid / liquid / gas. sent4: physical state means state of matter. sent5: shape is a kind of physical property. sent6: a physical property is a kind of property. sent7: neutrons are a kind of subatomic particle. sent8: best means most positive. sent9: the nucleus of an atom is made of protons and neutrons. sent10: good means positive. sent11: charge is a property of an object / a material / a substance and includes ordered values of negatively-charged / neutral / positively-charged. sent12: neutrons add mass to the nucleus. sent13: the nucleus of an atom is surrounded by  almost empty space. sent14: a structure of something is synonymous with a part of that something. sent15: the nucleus is the core of an atom. sent16: a neutron has a neutral electric charge. sent17: negative charge is a kind of electric charge. sent18: the location of an electron is determined by the attraction of the nucleus. sent19: \^+ means positive charge. sent20: nuclear fusion happens in the core of a star. sent21: mostly means being the most. sent22: a proton has a positive 1 electric charge. sent23: the sun is made of hydrogen and helium. sent24: the properties of something can be used to identify / used to describe that something. sent25: electron configuration is a property of elements.

\noindent
\textbf{Q:} How is the structure of an atom best described?

\noindent
\textbf{A:} The reasoning path is: sent16 \& sent22 -> int1: protons and neutrons together will be positive in charge; int1 \& sent9 -> int2: the nucleus of an atom is positive in charge; int2 \& sent15 -> int3: an atom has a positve charged core; sent13 \& sent15 -> int4: the core of an atom is surrounded by almost empty space; int3 \& int4 -> an atom has a positive core surrounded by mostly empty space. Therefore, the answer is a positive core surrounded by mostly empty space.

~\\
\textbf{C:} sent1: heating means adding heat. sent2: melting is a kind of phase change. sent3: state of matter is a property of matter and includes ordered values of solid / liquid / gas. sent4: mass has no impact on boiling / melting / freezing point. sent5: the ice melts. sent6: if the volume of a substance / of a matter expands while the mass remains the same, then the molecules in that substance / matter move farther apart. sent7: volume has no impact on boiling / melting / freezing point. sent8: melting point means temperature at which a solid melts / above which a solid melts. sent9: phase changes do not change mass. sent10: melting point is a property of a substance / material. sent11: constant means not changing. sent12: as a substance dissolves in water, the mass of the substance will stay the same. sent13: phase means state. sent14: the gravitational force of a planet / of a celestial object does not change the mass of an object on that planet or celestial body. sent15: ml means milliliters. 

\noindent
\textbf{Q:} A beaker containing 50 milliliters of ice is placed on a windowsill. After several hours, the ice melts. What property of the ice did not change when it melted?

\noindent
\textbf{A:} The reasoning path is: sent2 \& sent5 -> int1: the ice undergoes a phase change; int1 \& sent9 -> the mass of the ice will not change. Therefore, the answer is the mass.

~\\
\textbf{C:} sent1: gradual means slow / over time. sent2: amount is a property of something and includes ordered values of none / least / little / some / half / much / many / most / all. sent3: tremendous means intense. sent4: fast means quickly. sent5: tectonic plates being pushed together causes earthquakes. sent6: an earthquake wave is a kind of wave. sent7: to change means to cause a change. sent8: motion / movement means moving / to move. sent9: shaking the ground of the earth can change earth's surface. sent10: earthquake is a source of seismic waves. sent11: to cause means to result in. sent12: quickly means in a short period of time. sent13: a measure of time is a length of time. sent14: an earthquake causes the ground to shake. sent15: the ground is part of the earth. sent16: when an earthquake passes from the crust to the mantle, the wave of the earthquake changes speed. sent17: to happen means to occur. sent18: earthquakes are most common along tectonic plate boundaries. sent19: a surface is a part of an object. sent20: in a short amount of time is similar to quickly. sent21: earthquakes cause rock layers to fold on top of each other. sent22: an earthquake usually occurs over a period of 10 to 30 seconds. sent23: earthquake is a kind of vibration of the ground. sent24: the ground means earth's surface. sent25: 10 to 30 seconds is a short amount of time.

\noindent
\textbf{Q:} Natural processes cause rapid and slow changes to Earth's surface. Which event causes changes to Earth's surface most rapidly?

\noindent
\textbf{A:} The reasoning path is: sent22 \& sent25 -> int1: an earthquake usually occurs in a short amount of time; sent14 \& sent9 -> int2: an earthquake can change earth's surface by shaking the ground; int1 \& int2 -> int3: an earthquake can change earth's surface in a short amount of time; int3 \& sent20 -> an earthquake changes earth 's surface quickly. Therefore, the answer is the behavior is an earthquake shaking the ground.

~\\
\textbf{C:} sent1: sunlight means solar energy. sent2: the sun is the source of energy for life on earth. sent3: the sun is a source of radiation / heat called sunlight. sent4: solar power means solar energy. sent5: to provide something means to be the source of that something. sent6: the sun transfers solar energy / light energy / heat energy from itself to the planets / earth through sunlight. sent7: solar cells convert solar energy into electrical energy. sent8: billions of years is a longer period of time than 10 years. sent9: to provide means to supply. sent10: solar energy is a renewable / inexhaustible resource. sent11: photosynthesis means producers / green plants convert from carbon dioxide and water and solar energy into carbohydrates and food and oxygen for themselves. sent12: to depend on / to rely on / to need means to require. sent13: the sun is a kind of star. sent14: lifetime is similar to life cycle. sent15: the supply of an inexhaustible resource does not change over a long period of time. sent16: daylight means sunlight. sent17: a solar panel converts sunlight / solar energy into electrical energy / electricity. sent18: to change means to have an effect. sent19: to a human, a time period of 10 years or more years is considered very long. sent20: a measure of time is a length of time.

\noindent
\textbf{Q:} What makes solar energy different from most other energy that people use?

\noindent
\textbf{A:} The reasoning path is: sent10 \& sent15 -> int1: the supply of solar energy will not change over a long period of time; sent19 \& sent8 -> int2: to a human, a time period of billions of years is considered a very long time; int1 \& int2 -> the supply of solar energy will not change over billions of years. Therefore, the answer is the supply of solar energy will not change for billions of years.

~\\
\textbf{Non-sequential Reasoning}

~\\
\textbf{C:} sent1: a bear is a kind of animal. sent2: an animal is a kind of living thing. sent3: a forest is dark in color. sent4: protecting a living thing has a positive impact on that living thing 's survival / health. sent5: camouflage is a kind of adaptation for hiding in an environment. sent6: hiding is a kind of protection against predators / from predators. sent7: a forest is a kind of environment. sent8: a bear is dark in color. sent9: an example of camouflage is organism having the same color as its environment. sent10: if an animal 's fur is a color then that animal is that color. sent11: an animal is a kind of organism.

\noindent
\textbf{Q:} In Maryland, most bears have thick, dark fur. Which statement best explains how dark fur helps bears survive in the forests of Maryland?

\noindent
\textbf{A:} The reasoning path is: sent7 \& sent3 -> int1: a forest environment is dark in color; sent8 \& sent10 -> int2: a bear's fur is dark in color; int1 \& int2 -> int3: a bear's fur and a forest environment are both dark in color; sent1 \& sent11 -> int4: a bear is a kind of organism; int4 \& sent9 -> int5: an example of camouflage is a bear having the same color as its environment; int3 \& int5 -> int6: an example of camouflage is a bear's fur being the same dark color as a forest; int6 \& sent5 -> int7: a bear having dark fur is a kind of adaptation for hiding in a forest; sent4 \& sent2 -> int8: protecting an animal has a positive impact on that animal's survival; int8 \& sent6 -> int9: hiding can have a positive impact on an animal's survival; int9 \& sent1 -> int10: hiding can have a positive impact on a bear's survival; int10 \& int7 -> having dark fur has a positive impact on an animal's survival in a forest. Therefore, the answer is dark fur helps bears hide.

~\\
\textbf{C:} sent1: birds sometimes eat other animals. sent2: a predator eats other animals for food. sent3: prey is a kind of food for predators. sent4: claws are used to catch prey by some predators.

\noindent
\textbf{Q:} Which structure of a bird is correctly paired with its function?

\noindent
\textbf{A:} The reasoning path is: sent3 \& sent4 -> int1: claws are used by some predators to catch food; sent1 \& sent2 -> int2: some birds are predators; int1 \& int2 -> claws are used by some birds to catch food. Therefore, the answer is claws for obtaining food.

~\\
\textbf{C:} sent1: burning trees is used to produce heat. sent2: melting something off of something else is a lind of method for removing something from something else. sent3: wildfire is when a forest catches fire. sent4: something will perform an action after something else causes that something to be able to perform that action. sent5: the cones of a jack pine tree are serotinous cones. sent6: dispersing seeds is a kind of action for trees. sent7: removing something that prevents something else causes that something to be able to do that something else. sent8: a forest contains many trees. sent9: serotinous cones are sealed with a resin that prevents seed dispersal. sent10: resin can be melted with heat.

\noindent
\textbf{Q:} The cones of a jack pine tree require great amounts of heat to release their seeds. When are jack pine tree seeds most likely released?

\noindent
\textbf{A:} The reasoning path is: sent5 \& sent9 -> int1: the cones of a jack pine tree are sealed with a resin that prevents seed dispersal; int1 \& sent10 -> int2: the resin on the cones of a jack pine tree that prevent seed dispersal can be melted with heat; sent1 \& sent8 -> int3: burning a forest can produce great amounts of heat; int3 \& sent3 -> int4: a wildfire produces great amounts of heat; int2 \& int4 -> int5: a wildfire can melt the resin that prevents seed dispersal off of the cones of a jack pine tree; int5 \& sent2 -> int6: a wildfire can remove the resin that prevents seed dispersal from the cones of a jack pine tree; int6 \& sent7 -> int7: a wildfire causes jack pine tree cones to be able to disperse seeds; sent6 \& sent4 -> int8: a tree will disperse seeds after something else causes that tree to be able to disperse seeds; int7 \& int8 -> a jack pine tree will disperse seeds after a wildfire. Therefore, the answer is after a fire.

~\\
\textbf{C:} sent1: to take something in can mean to absorb something. sent2: a protein is a kind of nutrient. sent3: digestion is when an organism takes in nutrients from food into itself by eating. sent4: the digestive system digests food for the body. sent5: some foods contain proteins. sent6: the small intestine is a part of the digestive system. 

\noindent
\textbf{Q:} Each organ in the digestive system has a specific function. During digestion, food passes from the stomach into the small intestine. Which statement best describes what takes place in the small intestine?

\noindent
\textbf{A:} The reasoning path is: sent6 \& sent4 -> int1: digestion of food occurs in the small intestine; int1 \& sent5 -> int2: digestion of proteins occurs in the small intestine; int2 \& sent2 -> int3: digestion of nutrients occurs in the small intestine; sent1 \& sent3 -> int4: digestion is when an organism absorbs nutrients from food into itself; int3 \& int4 -> nutrients are absorbed from food in the small intestine. Therefore, the answer is nutrients are absorbed from the food.

~\\
\textbf{Least-to-Most Prompt (2-hop Exemplars)}

~\\
\underline{Decomposition Stage}

~\\
\textbf{Q:} If a place is in summer, then it will have the most sunlight. The northern hemisphere is a kind of hemisphere of Earth. A hemisphere of Earth is a kind of place. Melinda learned that days in some seasons have more daylight hours than in other seasons. Which season receives the most hours of sunlight in the Northern Hemisphere?

\noindent
\textbf{A:} To answer the question, we need to know: 'The Northern Hemisphere is a kind of what?', 'Which season receives the most hours of sunlight for this kind of thing?'.

~\\
\textbf{Q:} A dog is a kind of animal. Inheriting is when an inherited characteristic is copied / is passed from parent to offspring by genetics / dna. Fur is often part of an animal. The color of / coloration of fur is an inherited characteristic. Which of the following traits of a dog is inherited from its parents?

\noindent
\textbf{A:} To answer the question, we need to know: 'What does a dog often has?', 'Is this inherited from its parents?'.

~\\
\textbf{Q:} The Moon is the celestial object that is closer to the Earth than the Sun. The gravitational pull of the Moon on Earth 's oceans causes the tides. The gravitational pull of the Sun on Earth 's oceans causes the tides. As distance from an object decreases, the pull of gravity on that object increases. Tides on Earth are influenced more by the Moon than the Sun because the Moon is

\noindent
\textbf{A:} To answer the question, we need to know: 'What causes tides on the Earth?', 'What's the difference between this thing of the Moon and of the Sun?'.

~\\
\textbf{Q:} Adding heat to an object causes chemical reactions. Baking is when a human makes food by adding heat. A cake is a kind of food. Which of the following represents a chemical reaction?

\noindent
\textbf{A:} To answer the question, we need to know: 'What process includes chemical reactions?', 'What is an example of this process?'.

~\\
\underline{Solving Stage}

~\\
\textbf{C:} If a place is in summer, then it will have the most sunlight. The northern hemisphere is a kind of hemisphere of earth. A hemisphere of earth is a kind of place.

\noindent
\textbf{Q:} The Northern Hemisphere is a kind of what?

\noindent
\textbf{A:} The Northern Hemisphere is a kind of place.

\noindent
\textbf{Q:} Which season receives the most hours of sunlight for this kind of thing?

\noindent
\textbf{A:} A place receives the most sunlight in summer.

\noindent
Therefore, the answer is summer.

~\\
\textbf{C:} A dog is a kind of animal. Inheriting is when an inherited characteristic is copied / is passed from parent to offspring by genetics / dna. Fur is often part of an animal. The color of / coloration of fur is an inherited characteristic.

\noindent
\textbf{Q:} What does a dog often has?

\noindent
\textbf{A:} A dog often has fur since it is a kind of animal.

\noindent
\textbf{Q:} Is this inherited from its parents?

\noindent
\textbf{A:} The color of fur is an inherited characteristic from parents.

\noindent
Therefore, the answer is the color of its fur.

~\\
\textbf{C:} The Moon is the celestial object that is closer to the Earth than the Sun. The gravitational pull of the Moon on Earth 's oceans causes the tides. The gravitational pull of the Sun on Earth 's oceans causes the tides. As distance from an object decreases, the pull of gravity on that object increases.

\noindent
\textbf{Q:} What causes tides on the Earth?

\noindent
\textbf{A:} Tides on the Earth are caused by gravitational pull.

\noindent
\textbf{Q:} What's the difference between this thing of the Moon and of the Sun?

\noindent
\textbf{A:} The gravitational pull of the Moon on the Earth is larger than of the Sun because the Moon is closer to the Earth than the Sun.

\noindent
Therefore, the answer is closer than the Sun.

~\\
\textbf{C:} Adding heat to an object causes chemical reactions. Baking is when a human makes food by adding heat. A cake is a kind of food.

\noindent
\textbf{Q:} What process includes chemical reactions?

\noindent
\textbf{A:} Adding heat to an object includes chemical reactions.

\noindent
\textbf{Q:} What is an example of this process?

\noindent
\textbf{A:} Baking involves adding heat to food, and cake is a kind of food.

\noindent
Therefore, the answer is baking a cake.

\subsection{Prompts for QASC}

\textbf{Factual Prompt}

~\\
\textbf{C:} Decomposition increases the amount of nutrients in the soil. Bacteria and fungi are involved in the decomposition.

\noindent
\textbf{Q} What can increase the amount of nutrients in the soil? (A) roots (B) wind (C) worms (D) Fungi (E) Enzymes (F) corn (G) trees (H) mussels

\noindent
\textbf{A:} Fungi increase the amount of nutrients in the soil. Therefore, the answer is D.

~\\
\textbf{C:} An image in a mirror is formed by reflecting light. Water is essentially a perfect mirror, reflecting colors almost unchanged.

\noindent
\textbf{Q:} What forms an image in water? (A) dolphins (B) tiny polyps (C) dark matter (D) an object (E) reflecting light (F) individual molecules (G) holograms (H) jungles of the sea

\noindent
\textbf{A:} An image in water is formed by reflecting light. Therefore, the answer is E.

~\\
\textbf{C:} Pneumonia may be caused by an infection or injury of the lungs. Lung injuries are treated with bronchodilatory treatment.

\noindent
\textbf{Q:} What treatment can be used in treating pneumonia? (A) the respiratory system (B) a breathing mask (C) bronchodilatory treatment (D) aqueous solution (E) immunotherapy (F) radiation therapy (G) immune system (H) chemotherapy

\noindent
\textbf{A:} Pneumonia can be treated with bronchodilatory treatment. Therefore, the answer is C.

~\\
\textbf{C:} If an object reflects light toward the eye then that object can be seen. Light is focussed on the retina.

\noindent
\textbf{Q:} What does it mean if an object can be focussed on the retina? (A) sunlight (B) lowered energy cost (C) it has density (D) it is a rainbow (E) observe it (F) Change colors (G) it is invisble (H) it can be seen

\noindent
\textbf{A:} If an object can be focussed on the retina, it can be seen. Therefore, the answer is H.

~\\
\textbf{Counterfactual Prompt}

~\\
\textbf{C:} Decomposition increases the amount of nutrients in the soil. Enzymes are involved in the decomposition.

\noindent
\textbf{Q:} What can increase the amount of nutrients in the soil? (A) roots (B) wind (C) worms (D) Fungi (E) Enzymes (F) corn (G) trees (H) mussels

\noindent
\textbf{A:} Enzymes increase the amount of nutrients in the soil. Therefore, the answer is E.

~\\
\textbf{C:} An image in a mirror is formed by dark matter. Water is essentially a perfect mirror, reflecting colors almost unchanged.

\noindent
\textbf{Q:} What forms an image in water? (A) dolphins (B) tiny polyps (C) dark matter (D) an object (E) reflecting light (F) individual molecules (G) holograms (H) jungles of the sea

\noindent
\textbf{A:} An image in water is formed by dark matter. Therefore, the answer is C.

~\\
\textbf{C:} Pneumonia may be caused by an infection or injury of the lungs. Lung injuries are treated with radiation therapy.

\noindent
\textbf{Q:} What treatment can be used in treating pneumonia? (A) the respiratory system (B) a breathing mask (C) bronchodilatory treatment (D) aqueous solution (E) immunotherapy (F) radiation therapy (G) immune system (H) chemotherapy

\noindent
\textbf{A:} Pneumonia can be treated with radiation therapy. Therefore, the answer is F.

~\\
\textbf{C:} If an object reflects light toward the eye then that object has density. Light is focussed on the retina.

\noindent
\textbf{Q:} What does it mean if an object can be focussed on the retina? (A) sunlight (B) it has density (C) lowered energy cost (D) it is a rainbow (E) observe it (F) Change colors (G) it is invisble (H) it can be seen

\noindent
\textbf{A:} If an object can be focussed on the retina, it has density. Therefore, the answer is B.

\subsection{Prompts for bAbI15}

\textbf{Factual Prompt}

~\\
\textbf{C:} Jessica is a mouse. Sheep are afraid of wolves. Deer are afraid of lions. Winona is a wolf. Snakes are afraid of eagles. Emily is a deer. Gertrude is a eagle. Mice are afraid of cats.

\noindent
\textbf{Q:} What is Emily afraid of?

\noindent
\textbf{A:} Emily is a deer. We know that deer are afraid of lions. Therefore, Emily is afraid of lions.

~\\
\textbf{C:} Mice are afraid of cats. Gertrude is a mouse. Sheep are afraid of wolves. Winona is a cat. Snakes are afraid of eagles. Deer are afraid of lions. Emily is a sheep. Jessica is a eagle.

\noindent
\textbf{Q:} What is Gertrude afraid of?

\noindent
\textbf{A:} Gertrude is a mouse. We know that mice are afraid of cats. Therefore, Gertrude is afraid of cats.

~\\
\textbf{C:} Emily is a lion. Snakes are afraid of eagles. Gertrude is a sheep. Mice are afraid of cats. Winona is a cat. Sheep are afraid of wolves. Jessica is a wolf. Deer are afraid of lions.

\noindent
\textbf{Q:} What is Jessica afraid of?

\noindent
\textbf{A:} Jessica is a wolf. We only know that sheep are afraid of wolves. Therefore, what is Jessica afraid of is not mentioned.

~\\
\textbf{C:} Emily is a deer. Snakes are afraid of eagles. Gertrude is a wolf. Sheep are afraid of wolves. Deer are afraid of lions. Jessica is a snake. Winona is a lion. Mice are afraid of cats.

\noindent
\textbf{Q:} What is Gertrude afraid of?

\noindent
\textbf{A:} Gertrude is a wolf. We only know that sheep are afraid of wolves. Therefore, what is Gertrude afraid of is not mentioned.

~\\
\textbf{Counterfactual Prompt}

~\\
\textbf{C:} Cats are afraid of mice. Gertrude is a mouse. Wolves are afraid of sheep. Winona is a cat. Eagles are afraid of snakes. Lions are afraid of deer. Emily is a sheep. Jessica is a eagle.

\noindent
\textbf{Q:} What is Gertrude afraid of?

\noindent
\textbf{A:} Gertrude is a mouse. We only know that cats are afraid of mice. Therefore, what is Gertrude afraid of is not mentioned.

~\\
\textbf{C:} Jessica is a mouse. Wolves are afraid of sheep. Lions are afraid of deer. Winona is a wolf. Eagles are afraid of snakes. Emily is a deer. Gertrude is a eagle. Cats are afraid of mice.

\noindent
\textbf{Q:} What is Emily afraid of?

\noindent
\textbf{A:} Emily is a deer. We only know that lions are afraid of deer. Therefore, what is Emily afraid of is not mentioned.

~\\
\textbf{C:} Emily is a lion. Eagles are afraid of snakes. Gertrude is a sheep. Cats are afraid of mice. Winona is a cat. Wolves are afraid of sheep. Jessica is a wolf. Lions are afraid of deer.

\noindent
\textbf{Q:} What is Jessica afraid of?

\noindent
\textbf{A:} Jessica is a wolf. We know that wolves are afraid of sheep. Therefore, Jessica is afraid of sheep.

~\\
\textbf{C:} Emily is a deer. Eagles are afraid of snakes. Gertrude is a wolf. Wolves are afraid of sheep. Lions are afraid of deer. Jessica is a lion. Winona is an eagle. Cats are afraid of mice.

\noindent
\textbf{Q:} What is Winona afraid of?

\noindent
\textbf{A:} Winona is an eagle. We know that eagles are afraid of snakes. Therefore, Winona is afraid of snakes.

\end{document}